\documentclass[lettersize,journal]{IEEEtran}
\usepackage{amsmath,amsfonts}
\usepackage{algorithm}
\usepackage{algpseudocodex}
\usepackage{array}
\usepackage[caption=false,font=normalsize,labelfont=sf,textfont=sf]{subfig}
\usepackage{textcomp}
\usepackage{stfloats}
\usepackage{url}
\usepackage{verbatim}
\usepackage{graphicx}
\usepackage{cite}
\usepackage{hyperref}
\usepackage{colortbl}
\usepackage{xcolor}
\usepackage{makecell}
\usepackage{multirow}
\usepackage{cleveref}
\usepackage{booktabs}

\begin{document}

\newcommand{\fst}[1]{\textcolor{red}{\textbf{#1}}}
\newcommand{\snd}[1]{\textcolor{blue}{\textbf{#1}}}
\newcommand{\trd}[1]{\textbf{#1}}
\definecolor{MCyan}{HTML}{00FFFF}
\newcommand{\mzoom}[1]{\textcolor{red}{\textbf{#1}}}
\newcommand{\mdefect}[1]{\textcolor{MCyan}{\textbf{#1}}}
\newcommand{\nmtx}[1]{\mathbf{#1}}
\newcommand{\nscalar}[1]{\mathit{#1}}
\newcommand{\ncnst}[1]{\mathit{#1}}
\newcommand{\nvec}[1]{\mathbf{#1}}
\newcommand\norm[1]{\left\lVert#1\right\rVert}

\title{GSDeformer: Direct, Real-time and Extensible Cage-based Deformation for 3D Gaussian Splatting}

\author{Jiajun Huang, Shuolin Xu, Hongchuan Yu, and Tong-Yee Lee%
\thanks{This work was supported by HORIZON EUROPE Marie Sklodowska-Curie Actions, Grant 101130271, EPSRC Centre for Doctoral Training in Digital Entertainment, Grant EP/L016540/1, and the National Science and Technology Council, Taiwan, Grant 114-2221-E-006-114-MY3. \textit{(Corresponding authors: Hongchuan Yu, Tong-Yee Lee.)}}%
\thanks{Jiajun Huang, Shuolin Xu, and Hongchuan Yu are with the National Centre for Computer Animation, Bournemouth University, Poole BH12 5BB, U.K. (e-mail: jhuang@bournemouth.ac.uk; sxu@bournemouth.ac.uk; hyu@bournemouth.ac.uk).}%
\thanks{Tong-Yee Lee is with the Department of Computer Science and Information Engineering, National Cheng-Kung University, Tainan 70101, Taiwan. (e-mail: tonylee@mail.ncku.edu.tw).}}

\markboth{IEEE Transactions on Visualization and Computer Graphics,~Vol.~0, No.~0, January~2025}%
{Huang \MakeLowercase{\textit{et al.}}: GSDeformer: Direct, Real-Time, and Extensible Cage-Based Deformation for 3D Gaussian Splatting}


\maketitle

\begin{figure*}[!t]
\centering
\includegraphics[width=\textwidth]{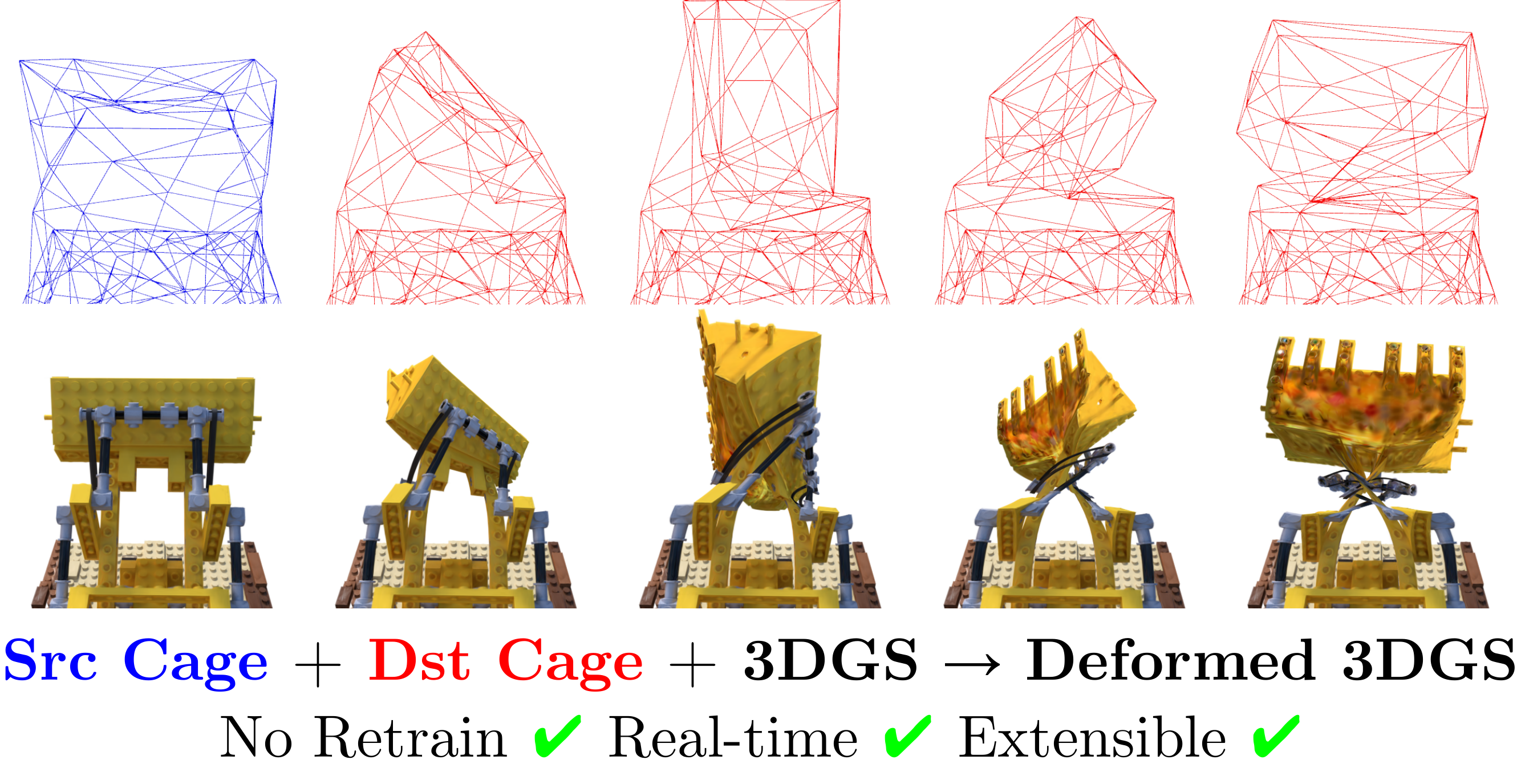}
\label{fig:teaser}
\end{figure*}

\begin{abstract}
We present GSDeformer, a method that enables cage-based deformation on 3D Gaussian Splatting (3DGS). Our approach bridges cage-based deformation and 3DGS by using a proxy point-cloud representation. This point cloud is generated from 3D Gaussians, and deformations applied to the point cloud are translated into transformations on the 3D Gaussians. To handle potential bending caused by deformation, we incorporate a splitting process to approximate it. Our method does not modify or extend the core architecture of 3D Gaussian Splatting, making it compatible with any trained vanilla 3DGS or its variants. Additionally, we automate cage construction for 3DGS and its variants using a render-and-reconstruct approach. Experiments demonstrate that GSDeformer delivers superior deformation results compared to existing methods, is robust under extreme deformations, requires no retraining for editing, runs in real-time, and can be extended to other 3DGS variants. Project Page:  \url{https://jhuangbu.github.io/gsdeformer/}
\end{abstract}

\begin{IEEEkeywords}
 3D Gaussian Splatting, Cage-based Deformation, 3D Representation Editing, Animation
\end{IEEEkeywords}
\section{Introduction}
3D Gaussian Splatting (3DGS) \cite{3dgs} is a novel and efficient approach for reconstructing and representing 3D scenes. Due to its ability to capture real-world objects and environments in impressive quality, it holds significant potential for downstream applications such as animation, virtual reality, and augmented reality. To make 3DGS practical for these applications, it is crucial to enable users to freely edit the captured scenes for privacy or creative purposes.

Current methods do not achieve direct, real-time, and extensible manipulation of 3D Gaussian Splatting (3DGS). Techniques like DeformingNeRF \cite{deforming-nerf}, CageNeRF \cite{cagenerf}, and NeRFShop \cite{nerf-shop}, which enable cage-based deformation on Neural Radiance Fields (NeRF), work by deforming sample points during the volumetric rendering process. However, since 3DGS does not use volumetric rendering, these methods cannot be easily adapted to it, restricting them to the more expensive-to-train and lower-quality NeRF-based representations.

While existing methods for editable 3DGS, such as SuGaR \cite{sugar} and Gaussian Frosting \cite{GaussianFrosting}, achieve high-fidelity deformation by tightly binding Gaussians to a high-resolution proxy mesh, they require significant modifications to the 3DGS representation or impose additional requirements on the training data, leading to the need for retraining. This limitation prevents these methods from directly editing existing 3DGS captures without extensive retraining, also making them harder to integrate with other 3DGS-derived scene representations. Similarly, methods like SC-GS \cite{SC-GS} and VR-GS \cite{vrgs} that allow manual editing rely on control structures such as tetrahedral grids or control point clouds, which are less convenient for integration with existing animation software and pipelines.

To address these challenges, we propose GSDeformer, a method that enables cage-based deformation on trained vanilla 3DGS models and their variants. Our approach operates directly on trained 3DGS without requiring extensive retraining, performs deformation in real time, and can be easily extended and integrated with other methods that enhance or build upon 3DGS.

Our approach leverages Cage-Based Deformation (CBD), which uses a coarse mesh (cage) to control deformations of the finer geometry within it. To deform the Gaussian distributions that make up the 3DGS representation, we generate a proxy point cloud from the Gaussians and apply CBD to deform the point cloud. The deformed proxy points then drive the transformations applied to the Gaussians. To handle potential bending caused by deformation, we introduce a splitting process for the relevant Gaussians.

This direct deformation approach enables editing of any trained vanilla 3DGS model without requiring architectural modifications or retraining. It also ensures compatibility with other 3DGS variants. Additionally, by using standard triangular cage meshes to control deformation, our method seamlessly integrates with existing animation software and pipelines.

Cages for deformation can be created either manually or automatically from 3DGS using our automated algorithm. Our automatic cage-building algorithm supports not only 3DGS but also its variants, such as 2DGS \cite{2dgs}, through a render-reconstruct-simplify approach. We also introduce a two-stage optimization design into our final mesh simplification process. This process is designed to produce a high-quality cage by both ensuring it better envelops the object to be deformed and improving the quality of the resulting MVC coordinates, which helps reduce artifacts in downstream deformation.

We evaluate the effectiveness of our method on object datasets. The results demonstrate that our deformation algorithm delivers superior quality compared to existing methods, particularly under extreme deformations. Among methods that enable explicit control, our control structure is the easiest to integrate with existing software. Furthermore, our algorithm achieves real-time performance ($\sim 60 FPS$), renders deformed representations at high speeds ($>200FPS$), and can be extended to other 3DGS variants.

In summary, our contributions include:

\begin{itemize}
    \item We propose GSDeformer, a method achieving real-time cage-based deformation on any trained 3D Gaussian Splatting(3DGS) model without re-training or altering its core architecture. 
    \item We also propose a robust automatic cage-generation algorithm that works on 3DGS as well as its variants due to its render-reconstruct-simplify approach. It ensures the creation of a high-quality cage by encouraging it to envelop the object better and produce stable Mean Value Coordinates through a two-stage process.
    \item We conduct extensive experiments to demonstrate our method's ease of integration with other work extending 3DGS, as well as showing our method's superior quality and ease-of-control against existing methods under normal and extreme scenarios.
\end{itemize}
  
\section{Related Work}

\subsection{Deformation of 3D Gaussian Splatting Scenes}

Many methods have been proposed to edit 3D Gaussian Splatting (3DGS) models. These range from high-level, textual-prompt-based editing methods such as GaussianEditor \cite{gaussianEditor}, VcEdit \cite{vcedit}, and GaussCtrl \cite{gaussCtrl}, to lower-level, more explicit editing methods that directly manipulate the scene's geometry.

To enable explicit editing, one effective approach is binding the Gaussian distributions in 3DGS to the surface of a dense proxy mesh surface. Deforming the 3DGS scene can then be achieved by deforming this mesh. This strategy allows for fine-grained, high-fidelity control over the geometry, as the high-resolution mesh provides a well-defined surface structure that can preserve intricate details during deformation. SuGaR \cite{sugar} pioneered this approach by extracting a mesh from a 3DGS scene, and GaussianFrosting \cite{GaussianFrosting} builds on it by proposing a more flexible way to bind distributions to the mesh. Other methods like GaMeS \cite{games} and Gao et al. \cite{gao-et-al} start with a provided initial mesh, while Mani-GS \cite{mani-gs} first trains a 3DGS or NeuS \cite{neus} model to create one. However, the trade-off for this high-fidelity control is that the binding process modifies the core 3DGS architecture to handle deformation, requiring costly retraining. This limitation makes it challenging to use these methods (\cite{sugar,GaussianFrosting,games,gao-et-al,mani-gs}) for modifying existing, pre-trained 3DGS scenes or extending them to new variants of the standard 3DGS. In contrast, our method prioritizes flexibility and compatibility by directly operating on the Gaussian primitives, eliminating the need for retraining or architectural changes and making it easier to integrate with variants of 3DGS.

In another direction, many works have explored physics-based simulation to manipulate 3DGS. PhysGaussian \cite{physgaussian} uses the Material Point Method (MPM) \cite{mpm14} to model object deformation from touch and push interactions on 3DGS, and Gaussian Splashing \cite{gsplashing} integrates position-based dynamics (PBD) \cite{PBD} with 3DGS for simulations. Similarly, VR-GS \cite{vrgs} employs a dense tetrahedral grid to enable real-time physical interaction. While the grid-based deformation mechanism bears some resemblance to cage-based approaches, its primary goal is simulating physical phenomena rather than providing the direct, manual manipulation central to our method. Overall, while these methods excel at replicating natural physics, our approach focuses on direct, user-controlled artistic deformation and manipulation.

Other work has explored direct manipulation of vanilla 3DGS models without requiring a dense proxy mesh. SC-GS \cite{SC-GS} enables Gaussian deformation by mapping control point transformations to Gaussians, but it requires video data for learning these mappings, limiting its applicability. D-MiSo \cite{dmiso}, similar to SC-GS, focuses on editing dynamic 3DGS scenes rather than static ones. Recent work ARAP-GS \cite{arapgs} adapts the As-Rigid-As-Possible deformation approach and resolves spike-like artifacts by aligning deformed Gaussians with their corresponding radiance fields. However, this method can be less effective in handling extreme deformations, such as the Gaussian highlighted in \Cref{fig:exp-ablation}, which shows the bending of a single Gaussian. Our observation is that splitting bent Gaussian ellipsoids into multiple smaller ones is essential. Our method addresses this limitation by introducing a novel splitting mechanism, which divides bent Gaussian ellipsoids into multiple smaller ones to better represent the deformed shape.

\subsection{Cage-Based Deformation}

Cage-based deformation (CBD) is a powerful and intuitive family of methods that use a coarse mesh, called a cage, to control the deformation of a more detailed inner mesh. CBD relies on cage coordinates to define how points within the cage relate to the cage's vertices. Various coordinate types, such as mean value coordinates (MVC) \cite{7mvc,18mvc}, harmonic coordinates (HC) \cite{5hc,17hc} and green coordinates (GC) \cite{22gc}, have been developed, all demonstrating strong performance in mesh deformation. Several methods, including DeformingNeRF \cite{deforming-nerf}, CageNeRF \cite{cagenerf}, NeRFShop \cite{nerf-shop}, Li et al. \cite{Li2023InteractiveGE}, and VolTeMorph \cite{voltemorph}, have adapted cage-based deformation for radiance fields. These approaches deform sample points along rays during volumetric rendering. However, this strategy is fundamentally incompatible with 3DGS, which uses rasterization, presenting a significant adaptation challenge.

The application of this powerful paradigm to 3DGS, however, remains under-explored. D3GA \cite{3dga} achieves cage-based deformation on 3DGS human bodies and garments by utilizing tetrahedral mesh cages. Tetrahedral meshes are effective for volumetric deformation, particularly for articulated subjects, but are less common in standard artistic modeling workflows than traditional triangular surface meshes. This leaves an opening for a general-purpose CBD framework that works on arbitrary 3DGS objects using standard, easy-to-edit triangular meshes, and can be extended to variants of 3DGS, which is the focus of this work.

A critical component of a practical CBD workflow is automatic cage construction. For NeRF, methods like NeRFShop \cite{nerf-shop} and DeformingNeRF \cite{deforming-nerf} employ marching cubes on the opacity field to create a fine mesh, which is then simplified. However, directly interpreting 3DGS as an opacity field is problematic due to noise and artifacts in trained models, particularly floaters. In contrast, our novel cage-generation algorithm adopts a render-reconstruct-simplify approach. By combining T-SDF integration, depth carving, and simplification based on Bounding Proxy \cite{bounding-proxy}, our method constructs high-quality cages not only for standard 3DGS models but also for surface-like variants (e.g., 2DGS \cite{2dgs} or Dai et al. \cite{dai2024surfels}), where marching cubes-based methods would fail, as they cannot be easily converted into opacity fields.

Finally, a key challenge in the mesh simplification process is ensuring the final cage both tightly encloses the object and maintains reasonable deformation coordinates. This requires maintaining non-negative Mean Value Coordinates (MVCs), as negative values can lead to artifacts. Previous methods have struggled to achieve both tightness and non-negative MVC coordinates simultaneously. Approaches like Neural Cage \cite{neural-cage} and CageNeRF \cite{cagenerf} enforce non-negativity by encouraging simple, convex cages, often at the cost of not being a tight fit that accurately captures the object's geometry. Conversely, methods that prioritize tightness, such as the original Bounding Proxy technique~\cite{bounding-proxy}, do not guarantee non-negative MVCs, potentially leading to deformation artifacts. Our approach resolves this dilemma by incorporating the non-negativity of MVCs as a novel two-stage optimization design within the QEM optimization during cage simplification. This ensures the final cage both tightly encloses the object and reduces deformation artifacts caused by negative coordinates.
\section{Method}


\subsection{Preliminaries}

\textbf{3D Gaussian Splatting} 3D Gaussian Splatting (3DGS) \cite{3dgs} is a method for representing 3D scenes using a set of 3D Gaussians. Each Gaussian is characterized by its mean $\nvec{\mu} \in \mathbb{R}^3$, covariance $\nmtx{\Sigma} \in \mathbb{M}^{3x3}$, opacity $\alpha \in \mathbb{R}$, and color parameters $\mathcal{P} \in \mathbb{R}^k$. The color is view-dependent, modeled using spherical harmonics with $k$ degrees of freedom. The covariance $\nmtx{\Sigma}$ is decomposed as $\nmtx{RSSR^\mathsf{T}\!}$, where $\nmtx{R}$ is a rotation matrix (encoded as a quaternion) and $\nmtx{S}$ is a scaling matrix (encoded as a scaling vector).

Our approach leverages 3DGS's key feature: representing scenes as a set of 3D Gaussians, each equivalent to an ellipsoid. This representation forms the foundation of our method.

\textbf{Cage-based Deformation} To deform a fine mesh using a cage, we consider a cage $\mathcal{C}_s$ with vertices $\{\nvec{v}_j\}$. Points $\nvec{x} \in \mathbb{R}^3$ inside the cage $\mathcal{C}_s$ can then be represented by cage coordinates $\{ \omega_j \}$ (e.g., mean value coordinates \cite{ju2005mean}). These coordinates define the position of $\nvec{x}$ relative to the cage vertices. The position of $\nvec{x}$ is calculated as the weighted sum of cage vertex positions:
\begin{align}
\nvec{x} = \sum_j \omega_j \nvec{v}_j
\end{align}

After deforming the cage from $\mathcal{C}_s$ to $\mathcal{C}_d$ with vertices $\{\nvec{v}'_j\}$, we can compute the new position $\nvec{x}'$ of $\nvec{x}$ using the calculated cage coordinates:

\begin{align}
\nvec{x}' = \sum_j \omega_j \nvec{v}'_{j}
\end{align}

The cage defines a continuous field within its boundaries, allowing it to deform the enclosed fine mesh by manipulating its vertices. This deformation process also works for arbitrary points within the source cage.

\textbf{Constrained Mesh Decimation} Mesh decimation aims to simplify a high-polygon mesh into a low-polygon one. A prominent technique to achieve this is iterative edge collapse, which simplifies the mesh by progressively removing edges from the mesh and merging their two endpoint vertices into a single new vertex. The Quadric Error Metrics (QEM) algorithm \cite{qem} is a popular method for guiding this process. It selects which edge to collapse and where to place the new vertex by minimizing a geometric error metric.

However, optimizing for error alone can produce artifacts like self-intersections. To prevent this, QEM can be augmented with constraints. For instance, Bounding Proxy \cite{bounding-proxy} uses a constrained QEM that formulates the placement of the new vertex $\nvec{v}$ for a collapsed edge $(i,j)$ as a convex quadratic program. The objective is to minimize the sum of squared distances to the planes of incident faces, subject to constraints that prevent local surface inversion:
\begin{equation}
\label{eq:qem}
\left\{
\begin{aligned}
& \min_{\nvec{v}} E_{ij}(\nvec{v}) = \nvec{v}^T \nmtx{Q} \nvec{v} \\
& \text{s.t. } D(\nvec{v}, f) = \nvec{n}_f^T(\nvec{v} - \nvec{p}_f) \ge 0, \forall f \in \text{EU}(i, j)
\end{aligned}
\right.
\end{equation}
Here, $\nmtx{Q}$ is the sum of incident plane quadrics, and $\text{EU}(i,j)$ is the set of faces in the 1-ring neighborhood of the edge. The constraint $D(\nvec{v},f)$ ensures the new vertex does not move to the wrong side of its neighboring faces, where $\nvec{n}_f$ is the outward normal of face $f$.

\subsection{Cage-Building Algorithm}

\begin{figure*}
  \centering
  \includegraphics[width=\linewidth]{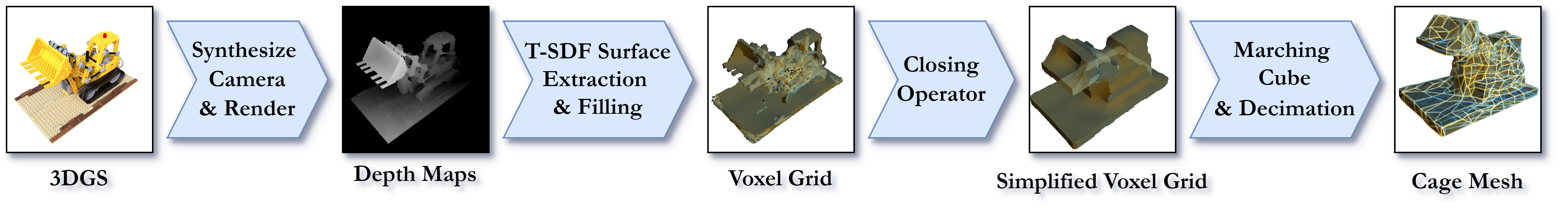}
  \caption{Overview of our cage-building algorithm. Given an object, our method renders a depth image from it, performs T-SDF integration, surface extraction, and space carving to produce a solid voxel grid. The voxel grid is then simplified using a morphological closing operator, meshed using marching cubes, and decimated to obtain the final cage mesh.}
  \label{fig:method-cage-building}
\end{figure*}

Given a trained 3DGS (or its variants) model $\mathcal{S}_s$, our automated cage-generation algorithm aims to create a simple cage $\mathcal{C}_s$ that encloses it. 

The proposed algorithm revolves around a more robust "render-reconstruct-simplify" pipeline, as illustrated in Figure \ref{fig:method-cage-building}. Most notably, instead of interpreting the 3DGS scene directly as an opacity field, our method extracts the underlying geometry by rendering 2D depth maps from multiple views. Additionally, our final decimation process employs a two-stage process that encourages the non-negativity of internal MVC coordinates, ensuring stable and artifact-free deformations. The full process is broken down into the following steps, following Figure \ref{fig:method-cage-building}:

\textbf{Synthesize Camera and Render} The algorithm begins by synthesizing a set of virtual cameras arranged on a sphere that encompasses the 3DGS scene. Depth maps are then rendered from each of these camera positions. This initial rendering step serves two crucial purposes. First, it aligns better with the native, rendering-based nature of 3DGS, making the geometry extraction less susceptible to artifacts like low-opacity floaters than methods that treat the scene as an opacity field. Second, by converting the scene into a standardized format of depth maps, the pipeline becomes agnostic to the specifics of the underlying 3DGS representation, ensuring its applicability to both standard 3DGS and its non-volumetric variants.

\textbf{T-SDF Surface Extraction and Filling} The collection of rendered depth maps is then fused into a single, coherent 3D volume using the Truncated Signed Distance Field (T-SDF) integration algorithm from KinectFusion \cite{kinectfusion}. By integrating depth information from multiple viewpoints, the T-SDF reconstruction naturally averages out inconsistencies and suppresses noise caused by render errors or floaters, resulting in a significantly cleaner surface representation. 

To produce a solid volume, this process is followed by a space-carving step. Clean depth maps are re-rendered from the newly created T-SDF volume and used to carve away empty space, filling the interior of the surface. This process yields a high-resolution, solid voxel grid that represents the object's geometry without the artifacts present in the original 3DGS scene.

\textbf{Apply Morphological Closing Operator} While the reconstructed voxel grid is geometrically accurate, its high level of detail is unsuitable for a deformation cage. To generate a simplified, coarse envelope, a morphological closing operator, as described in Bounding Proxy \cite{bounding-proxy}, is applied. This operator is a combination of dilation and erosion, aiming to remove noise and details from the voxel grid, producing a simplified, solid volumetric shape that smoothly envelops the original geometry, making it an ideal precursor to the final cage.


\textbf{Marching Cube and Decimation} In the final stage, we first apply the Marching Cubes algorithm to the simplified voxel grid to extract a dense polygonal mesh. This mesh is then simplified using a constrained edge-collapse algorithm \cite{bounding-proxy} to produce the final low-polygon cage.

Although the standard constrained QEM used in Bounding Proxy~\cite{bounding-proxy} produces a cage that tightly encloses the object, many object vertices may still lie outside the cage. This increases the occurrence of negative MVC weights, which in turn causes artifacts in MVC-based deformation. To produce a cage that provides good shape coverage and is optimized for high-quality deformations, our method adopts an alternating two-stage decimation procedure.

Stage 1: Constrained QEM. First, we perform edge collapse using the standard constrained QEM described in the preliminaries. Each step removes one edge by merging its two endpoints, while strictly enforcing constraints that prevent local surface inversion and ensure the cage remains manifold.

Stage 2: Refinement with MVC Penalty. After a fixed number of collapses, we refine the positions of all cage vertices to reduce negative MVC weights. Specifically, we minimize the following loss:
$$
L_{\text{MVC}}(V) = \frac{\mu}{|V|\,|S|}\sum_{i \in S}\sum_{\nvec{v}_j \in V} \bigl|\min\bigl(\phi_{ji}(V),\, 0\bigr)\bigr|^2 + \rho\,\|V - V_0\|^2
$$
where $V = \{\nvec{v}_j\}$ is the set of all cage vertex positions, $\phi_{ji}(V)$ is the mean value coordinate of a sample point $i \in S$ on the object surface with respect to cage vertex $\nvec{v}_j$, and $V_0$ denotes the positions before refinement. The coefficient $\mu \ge 0$ controls the strength of the MVC penalty, while the regularization term, weighted by $\rho \ge 0$, anchors the vertices near their pre-refinement positions to stabilize the optimization. We minimize $L_{\text{MVC}}$ via gradient descent for a fixed number of iterations.

We first run Stage~1 until the vertex count falls below a prescribed threshold, then alternate between the two stages: Stage~1 collapses a number of edges, then Stage~2 optimizes all vertex positions for a number of gradient-descent steps, and so on. This interleaving ensures that the cage remains geometrically valid throughout decimation while progressively improving shape coverage and minimizing negative MVC weights, yielding a cage that is structurally suited for stable, high-quality deformations.

In summary, our method creates high-quality deformation cages through a robust geometry extraction stage followed by a deformation-aware simplification process. By utilizing depth maps and T-SDF integration, we first reconstruct a clean representation of the object, which effectively bypasses 3DGS artifacts like floaters and supports non-volumetric variants of 3DGS. When converting it into a coarse cage, our novel two-stage decimation algorithm optimizes the cage structure. This two-stage process represents a deliberate trade-off: while concave regions necessary for tight enclosure are preserved, the application of the MVC penalty drives negative MVC weights towards zero (see \Cref{tbl:neg-mvc-quant}), resulting in a deformation cage that is not merely a geometric envelope but is structurally optimized for stable, high-fidelity deformation with significantly reduced artifacts.


\begin{figure*}[t]
  \centering
  \includegraphics[width=\textwidth]{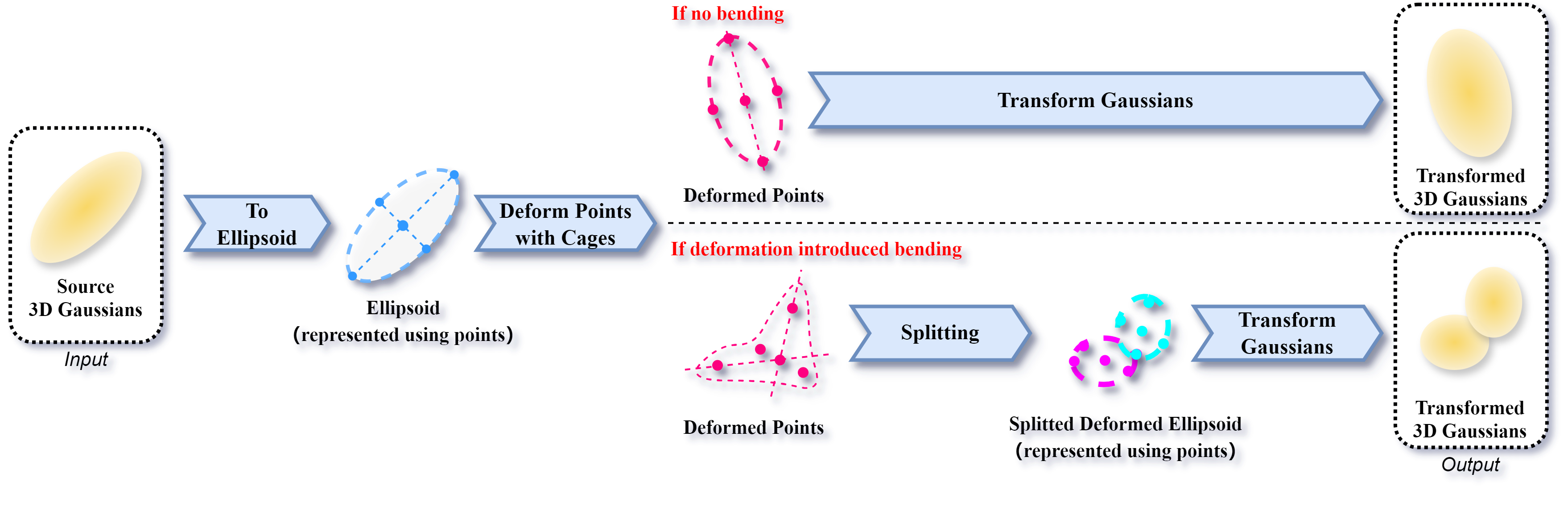}
  \vspace{-1em}
  \caption{Overview of our deformation algorithm. The deformation process is shown in 2D for clarity. For deformation, 3DGS Gaussians are converted to ellipsoids represented using points (the proxy point cloud). Proxy points are deformed using cage-based deformation and split if their axes are bent. Finally, deformed points are used to infer transformations for the Gaussians. For more details, please refer to the pseudo-code in Appendix \ref{sect:apptx-pseudocode}.}
  \label{fig:method}
  \vspace{-1em}
\end{figure*}

\subsection{Deformation Algorithm}


Our deformation algorithm takes a trained 3DGS scene $\mathcal{S}_s$ along with source and target cages, $\mathcal{C}_s$ and $\mathcal{C}_d$. These cages define a deformation for part or all of the scene. The objective is to produce the deformed scene $\mathcal{S}_d$, a 3DGS representation with the specified deformation applied.

The algorithm performs deformation on the 3D Gaussians that constitute the 3DGS scene representation. For each Gaussian $s$ with mean $\nvec{\mu}_{s} \in \mathbb{R}^3$ and covariance $\nmtx{\Sigma}_s \in \mathbb{M}^{3x3}$ (encoded by a rotation matrix $\nmtx{R}$ and scaling matrix $\nmtx{S}$), our deformation process is applied. This process is illustrated in \Cref{fig:method}. For a detailed algorithmic description, please refer to the pseudocode provided in Appendix \ref{sect:apptx-pseudocode}.

\textbf{To Ellipsoid} To geometrically manipulate the 3D Gaussians, we start by converting them into explicit geometries, namely ellipsoids. The ellipsoids are represented using point clouds so they can be deformed by cage-based deformation.

According to 3DGS \cite{3dgs}, a 3D Gaussian with mean $\nvec{\mu}_s$ and covariance $\nmtx{\Sigma}_s$ is defined as:
$$
G(x) = \exp\left\{-\frac{1}{2}(\nvec{x}-\nvec{\mu}_s)^\mathsf{T}\!\nmtx{\Sigma}_s^{-1}(\nvec{x}-\nvec{\mu}_s)\right\}
$$

The ellipsoid is then defined in quadric form as:
$$
(\nvec{x}-\nvec{\mu}_s)^\mathsf{T}\!\nmtx{\Sigma}_s^{-1}(\nvec{x}-\nvec{\mu}_s) = 1
$$

Here, $\nvec{\mu}_s$ is the ellipsoid's center $\nvec{c}$. The principal axes and their lengths are derived from the eigenvalue decomposition of $\nmtx{\Sigma}_s^{-1}$, that is;

\begin{itemize}
\item The three principal axes (e.g., $x$, $y$, $z$) correspond to the eigenvectors.
\item The axis lengths are inversely proportional to the eigenvalues of $\nmtx{\Sigma}_s^{-1}$.
\end{itemize}

We define six intersection points of the ellipsoid's principal axes with its surface as:
\begin{equation}
\label{eq:point-set}
AP = \{\nvec{c}, \nvec{x_1}, \nvec{y_1}, \nvec{z_1}, \nvec{x_2}, \nvec{y_2}, \nvec{z_2}\}
\end{equation}

where $\nvec{c}$ is the center of the ellipsoid, $\nvec{x}_{1,2}$ denote the two intersection points on the $x$-axis, with similar definitions for $y$ and $z$. These points, $AP$, describe the ellipsoid and form a proxy point cloud for cage-based deformation.

\textbf{Deform Points with Cages} With the proxy point cloud $AP_s$ in place, we apply the desired deformation defined by the source cage $\mathcal{C}_s$ and target cage $\mathcal{C}_d$ to it.

More concretely, we transform $AP_s$ using Mean Value Coordinates (MVC) \cite{7mvc}. First, we convert $AP_s$ from Euclidean coordinates to MVC using $\mathcal{C}_s$. Then, we convert them back to Euclidean coordinates using $\mathcal{C}_d$. The resulting deformed axis points are denoted as ${AP}_d$.

\textbf{Transform Gaussians} With the original and deformed points, we then estimate the underlying transformation and apply it to the original 3D Gaussians $s$. 

Given the original and deformed points $AP_s$ and $AP_d$, we compute the transformation $\nmtx{T} \in \mathbb{R}^{3\times3}$ by minimizing:
\begin{equation}
\min_\nmtx{T} \|\nmtx{D}_d - \nmtx{T}\nmtx{D}_s\|^2
\end{equation}

where $\nmtx{D} = [ \nvec{x_1} - \nvec{c}, \nvec{y_1} - \nvec{c}, \nvec{z_1} - \nvec{c}, \nvec{x_2} - \nvec{c}, \nvec{y_2} - \nvec{c}, \nvec{z_2} - \nvec{c} ] \in \mathbb{R}^{3\times6}$ for $AP_s$ or $AP_d$. The least squares solution for T is given by:
\begin{equation}
\label{eq:estimate-transform}
\nmtx{T} = \nmtx{D}_d\nmtx{D}_s^\mathsf{T}\!(\nmtx{D}_s\nmtx{D}_s^\mathsf{T}\!)^{-1}
\end{equation}

The 3D Gaussian $s$ can then be transformed using $\nmtx{T}$, while using the center of deformed points $AP_d$ as the new mean:
\begin{gather}
\label{eq:apply-transform}
\nvec{\mu}_d = \nvec{c}_d \\
\nmtx{\Sigma}_d = \nmtx{T} \nmtx{\Sigma}_s \nmtx{T}^\mathsf{T}\!
\label{eq:apply-transform-1}
\end{gather}

This step is crucial to our algorithm as it updates the Gaussians' covariance, which models their shape and orientation. Not updating covariance would lead to severe artifacts, which we will demonstrate in ablation studies.

According to the implementation of 3DGS \cite{3dgs}, the transformed mean and covariance can be directly used for rendering. Optionally, to recover the rotation and scaling parameters compatible with 3DGS, we re-fit the potentially sheared covariance matrix $\nmtx{\Sigma}_d$ via eigendecomposition: $\nmtx{\Sigma}_d = \nmtx{R}_d \nmtx{\Lambda}_d \nmtx{R}_d^\mathsf{T}\!$. The new rotation is the orthogonal matrix of eigenvectors $\nmtx{R}_d$, and the new scaling is computed from the square roots of the eigenvalues in $\nmtx{\Lambda}_d$.

\begin{figure}
  \centering
  \includegraphics[width=\linewidth]{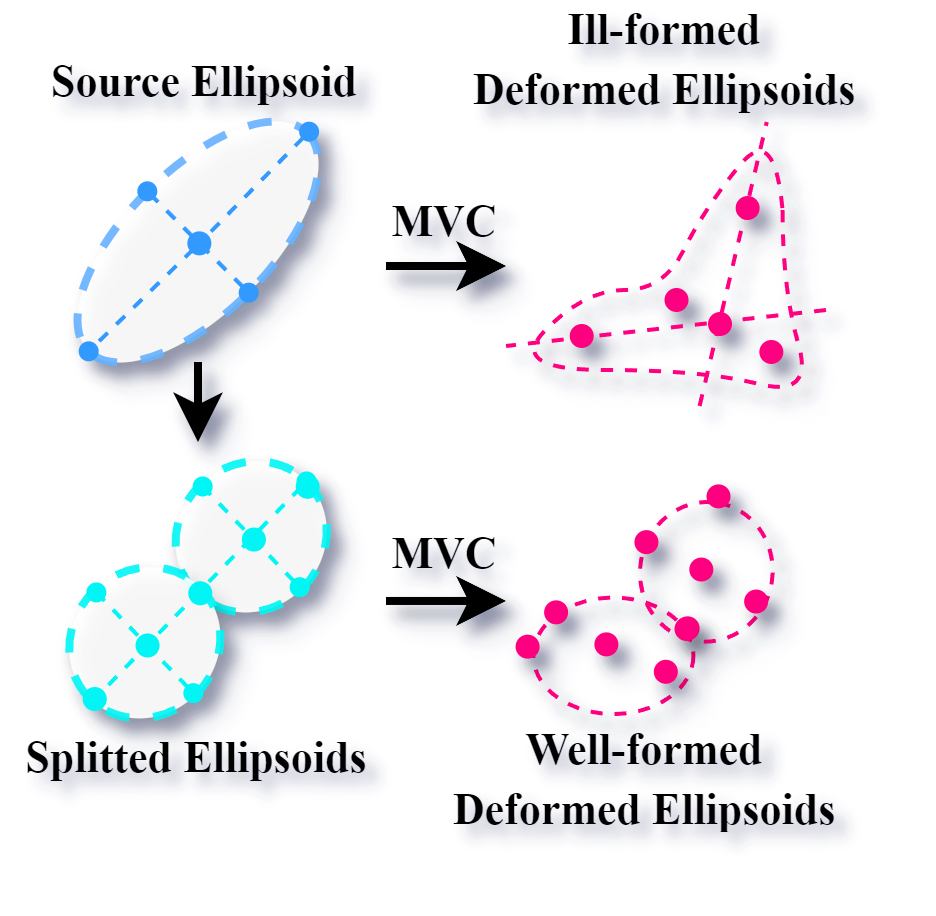}
  \caption{The splitting process. Our method fixes the ill-formed bent Gaussian by splitting the Gaussian before MVC deformation, leading to well-formed Gaussians and, thus, reasonable transforms.}
  \label{fig:method-splitting}
\end{figure}

\textbf{Splitting} Cage-based deformation enables flexible shape manipulation but introduces non-rigid warping, requiring some Gaussians to bend in deformation. This leads to visual artifacts since a single transformed Gaussian cannot properly represent these bent shapes. To address this, we split the bent Gaussians into multiple well-formed Gaussians to approximate the bent shapes, as shown in \Cref{fig:method-splitting}.

We start by identifying the bent Gaussians to split. Given a Gaussian with deformed points ${AP}_d$, its center is $\nvec{c}'_d \in {AP}_d$. The deformed points of an axis (e.g., x-axis) are $\nvec{x}'_{1}, \nvec{x}'_{2} \in {AP}_d$. We calculate the angle between vectors, $\nvec{x}'_{1} - \nvec{c}'$ and $\nvec{x}'_{2} - \nvec{c}'$. Splitting is required if the angle falls below a threshold.

The splitting method divides a pre-MVC Gaussian ellipsoid into two smaller ellipsoids along a chosen axis. Each new ellipsoid has its axis lengths reduced by factor $k$, and both are positioned to meet at the center of the original ellipsoid along the splitting axis. Finally, both resulting ellipsoids undergo MVC transformation again for further processing. The resulting two smaller ellipsoids are more tolerable to deformation along that axis, reducing the risk of bending or distortion when applying MVC directly to elongated shapes. This identification and splitting process is run across all axes to ensure that Gaussians requiring splitting on multiple axes are properly split more than once.

We determine the axis scaling factor $k$ by analyzing the optimal value that can maximally preserve the volume and shape of the original ellipsoid. More concretely, our scaling factor should satisfy two key constraints: Volume Preservation and Shape Preservation.

Volume Preservation Constraint: The volume of the two smaller ellipsoids combined should approximate the volume of the original ellipsoid. This can be expressed as: 
\begin{equation}
\label{equ:split-loss}
L_V = \frac{2}{3}\pi abc - 2\frac{2}{3}\pi k_1 ak_2 bk_3 c
\end{equation}

where, $a$, $b$, $c$ are the semi-principal axes of the original ellipsoid, $k_1$, $k_2$, $k_3$ are the individual scaling factors for the smaller ellipsoids. Simplifying this equation leads to: 
\begin{equation}
k_1k_2k_3 = \frac{1}{2}
\end{equation}

Shape Preservation Constraint: Each smaller ellipsoid should maintain a shape similar to the original ellipsoid. This constraint can be expressed as: 
\begin{equation}
\label{equ:split-equal}
\frac{k_1a}{k_2b} \approx \frac{a}{b} \quad \frac{k_2b}{k_3c} \approx \frac{b}{c} \quad \frac{k_1a}{k_3c} \approx \frac{a}{c}
\end{equation}

From these constraints, the scaling factor for all axes becomes:
\begin{equation}
    k_1 = k_2 = k_3 = \sqrt[3]{\frac{1}{2}}
\end{equation}

Thus, the scaling factor $k$ for splitting should be $\sqrt[3]{\frac{1}{2}}$. 

In practice, setting $k = \frac{1}{2}$ produces marginally better results, as demonstrated by our ablation study.

In summary, our algorithm deforms 3DGS by first converting 3D Gaussians into ellipsoids represented using points. Deformations of the points can then be transferred back to the underlying 3D Gaussians. This approach directly operates on the primitives of 3DGS, thus no architectural changes or retraining is needed. Furthermore, with caching, this direct transform approach can work in real-time. Finally, since it operates directly on the primitives, our method can be easily extended to work with 3DGS variants without concerns on training (such as FLoD \cite{flod}) and integrate with other 3DGS editing work (such as GaussianEditor \cite{gaussianEditor}) with minimal changes.

\textbf{Remark} We propose a splitting procedure to handle non-rigid warping that bends Gaussians. This approach improves deformation quality for scenes with intricate details, especially when a single Gaussian representing a straight surface must be bent. Unfortunately, the existing methods, such as ARAP-GS\cite{arapgs} and VR-GS \cite{vrgs}, ignore this issue.

\section{Experiments}

\subsection{Implementation Details}

For cage building, the voxel grid's resolution is set to 128. We employ the T-SDF integration and carving procedure from Open3D \cite{open3d}, using their default parameters. For the alternating decimation, in each cycle Stage~1 collapses $10$ edges, and Stage~2 runs for $10$ gradient-descent steps. We begin alternating at 1000 vertices, sample $|S|=256$ points on the object surface, and set $\mu=100$, $\rho=10^{-4}$. The MVC refinement stage uses the Adam optimizer with a learning rate of $0.005$.

For deformation, we set the splitting threshold at 175 degrees and avoid splitting axes with lengths below 1e-2 for numerical stability. For cages that only encompass part of the scene, we compute the convex hull of the cage and only deform gaussians fall inside it for speed and stability.

We ran all experiments using an NVIDIA A5000 GPU and an AMD Ryzen Threadripper PRO 3975WX processor (32 cores).

\subsection{Cage Building Quality}

\begin{figure*}
  \centering
  \includegraphics[width=\textwidth]{figures/exp-qual-caging}
  \caption{Comparison of cage building algorithm. We present the raw voxel grids and the produced final cages for comparison. \mzoom{Red boxes} indicate zoomed areas; \mdefect{cyan circles} marks defects. Note that our method generates less noisy voxel grids and floater/cavity-free final cages for both 3DGS and 2DGS. The marching cube-based baseline suffers from floaters or hollow cavities, especially on 2DGS, a non-volumetric variant of 3DGS. }
  \label{fig:exp-qual-caging}
\end{figure*}

\begin{table} 
    \centering
    \resizebox{\linewidth}{!}{
        \Huge
        \begin{tabular}{cc||*{2}{c}}
    \toprule Scene & Method & Mesh Components($\downarrow$) & Self-intersections($\downarrow$) \\
          
    \midrule \multirow{2}{*}{Lego} & Baseline & 2 & 0 \\ 
     & Ours & \fst{1} & \fst{0} \\ 
    
    \midrule \multirow{2}{*}{Vase} & Baseline & 2 & 3 \\ 
     & Ours & \fst{1} & \fst{0} \\ 

    \midrule \multirow{2}{*}{Vase (2DGS)} & Baseline & 3 & 5 \\ 
     & Ours & \fst{1} & \fst{0} \\ 

    \bottomrule
\end{tabular}
    }
    \vspace{0.5em}
       \caption{Quantitative comparison on the quality of the generated cage. Note that the baseline suffers from multiple mesh components (e.g., floaters or hollow cavities) and self-intersections, whereas our method does not.}
   \label{tbl:caging-quant}
    \vspace{-3.5em}
\end{table}

Our evaluation begins by analyzing the effectiveness and extensibility of our cage-building algorithm.

We evaluate our approach against the traditional marching-cube-then-simplify process used in NeRFShop \cite{nerf-shop} and DeformingNeRF \cite{deforming-nerf}. However, since the process used by existing methods is designed for radiance fields and cannot directly work with 3DGS, we adapt them based on SuGaR's \cite{sugar} approach, as detailed in Appendix \ref{sect:cage-building-baseline}.

We evaluate both methods using the Lego scene from the NeRF Synthetic Dataset \cite{nerf} and a flower vase extracted from the MipNeRF360 Dataset's garden scene \cite{mipnerf360}. To quantitatively evaluate the quality of the generated cages, we also computed their mesh components and the number of intersecting face pairs to detect floaters, hollow cavities, and self-intersections.

In \Cref{fig:exp-qual-caging}, we compare the output of our method with the marching cube-based baseline, showing both the raw voxel grids (before any smoothing operation) and the final meshes. Our method yields cleaner voxel grids for Lego and vase scenes, leading to higher-quality cages. In contrast, the baseline's voxel grids are significantly noisier and contain more floaters, resulting in self-intersections and floaters in the Lego and vase cages, as shown by \Cref{tbl:caging-quant}.

Our cage-building algorithm also extends naturally to 3DGS variants. When applied to a trained 2DGS representation of the vase (third row of \Cref{fig:exp-qual-caging} and \Cref{tbl:caging-quant}), the MC baseline method fails to produce a coherent voxel grid before any smoothing, leading to cages with hollow cavities and floaters, while our approach still generates accurate voxel grids and cages suitable for cage-based deformation. This difference occurs because 2DGS uses flat 2D ellipses with minimal volume instead of volumetric 3D ellipsoids. Our method remains effective regardless of this distinction.

\subsection{Deformation Quality}

\begin{figure*}
  \centering
  \includegraphics[width=\textwidth]{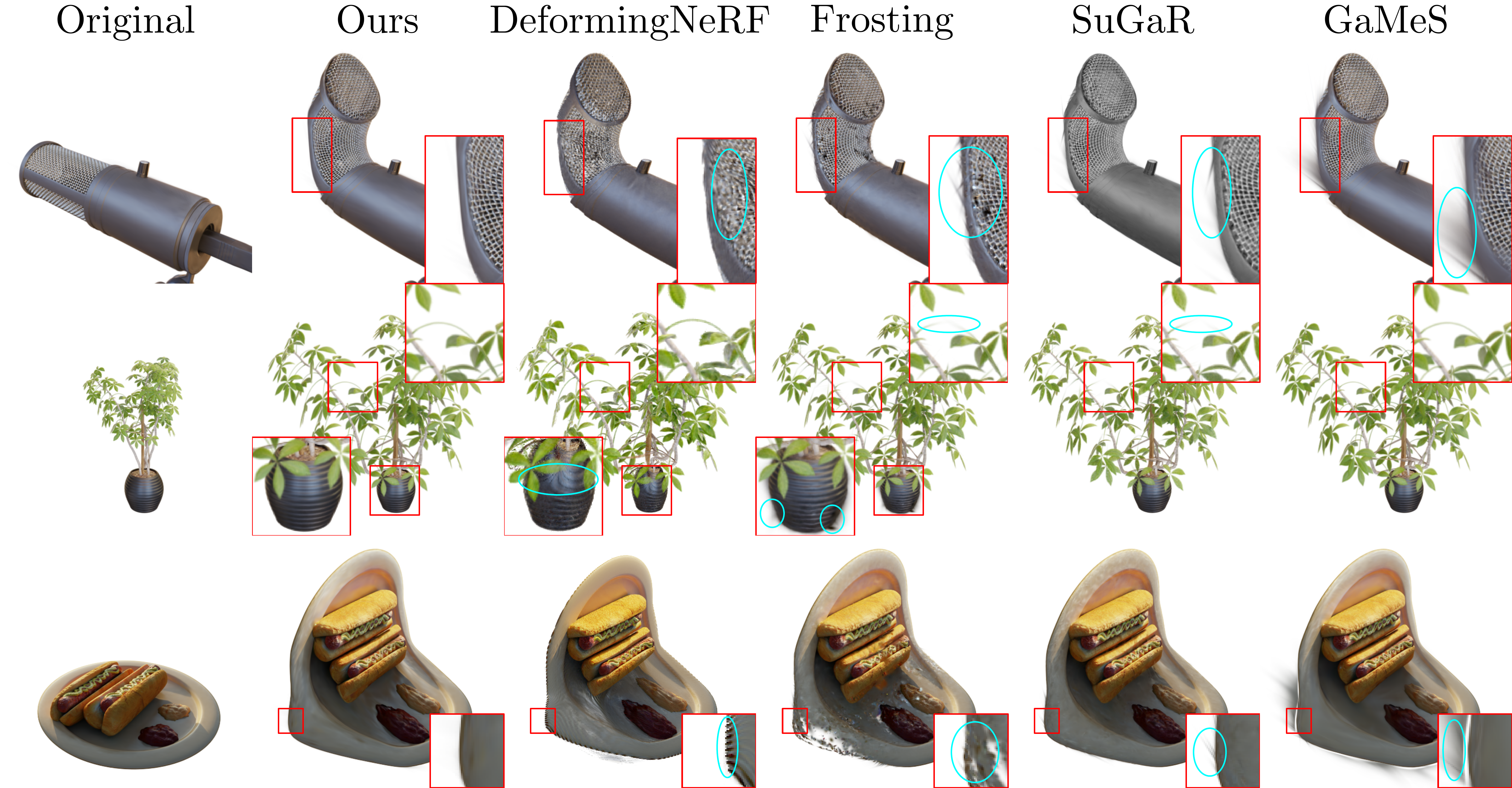}
  \caption{Comparison of methods on selected objects. All methods shown are driven by the exact same cage for a fair comparison. \mzoom{Red boxes} indicate zoomed areas; \mdefect{cyan circles} marks defects. Not having defect marks indicates satisfactory results. Our approach is the only method that performs well across all cases. For more results, refer to our supplementary video.}
  \label{fig:exp-qual}
\end{figure*}

\begin{figure*}
  \centering
  \includegraphics[width=\textwidth]{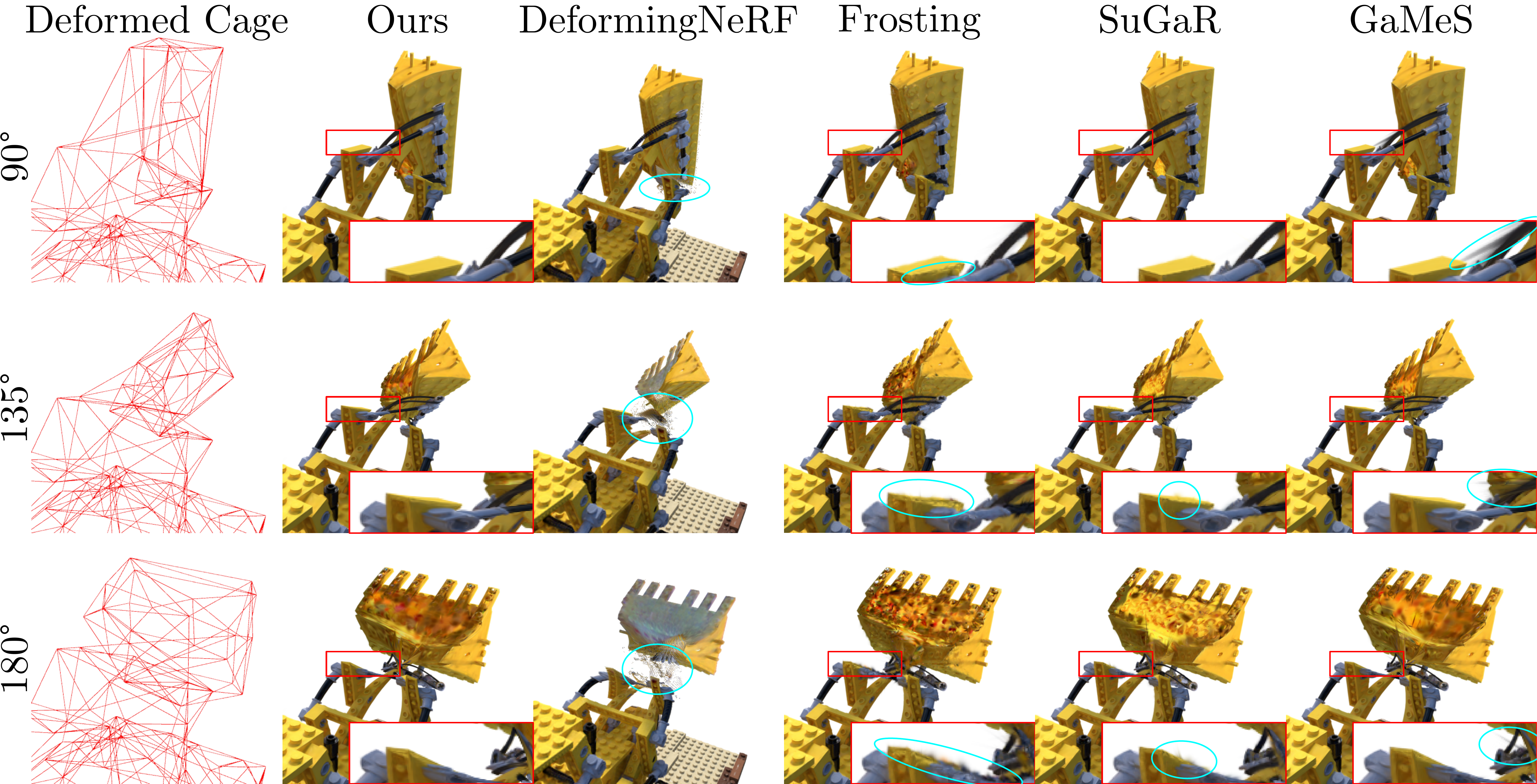}
  \caption{Comparison of methods from normal to extreme deformations. All methods shown are driven by the exact same cage (shown on left) for a fair comparison. \mzoom{Red boxes} indicate zoomed areas; \mdefect{cyan circles} marks defects. Not having defect marks indicates satisfactory results. Note that our method remains robust as deformation intensifies, while other methods develop artifacts. Even under the 180-degree extreme scenario, our method still produces reasonable results.}
  \label{fig:exp-qual-interp}
\end{figure*}

We then compare the deformation quality of our model against existing methods on the NeRF Synthetic Dataset \cite{nerf}. We start with pre-trained 3D Gaussian Splatting \cite{3dgs} models, apply our cage construction algorithm for cages, manually deform the cages, and run our deformation algorithm. To ensure a direct and fair comparison, we apply the identical deformed cage to all baseline methods. Specifically, for approaches like DeformingNeRF \cite{deforming-nerf}, GaMeS \cite{games}, SuGaR \cite{sugar}, and Gaussian Frosting \cite{GaussianFrosting}, we use our cage to deform their underlying mesh or triangle soup, ensuring that any differences in output quality are attributable to the deformation algorithms themselves and not the control input.

\textbf{Normal Deformations} We present our results in \Cref{fig:exp-qual}. In the microphone scene, DeformingNeRF fails to preserve the detailed grid structure of the microphone's mesh, while Gaussian Frosting produces holes and spiky artifacts on it. GaMeS and SuGaR also show spiky artifacts. For the ficus scene, DeformingNeRF and Gaussian Frosting create artifacts on the unedited flower pot. In the expanded upper part, Gaussian Frosting and SuGaR struggle with details, breaking connections between branches. In the hotdog scene, there are wrinkles on the plate with DeformingNeRF, severe tearing with Gaussian Frosting, and spiking artifacts with SuGaR and GaMeS due to the lack of a splitting process. Across all scenes, our approach is the only method that consistently produces the smoothest and most plausible results.

\textbf{Extreme Deformations} We further test how well these methods handle challenging deformations. In \Cref{fig:exp-qual-interp}, we rotate the bulldozer's head in the Lego scene from the NeRF Synthetic Dataset by 90, 135, and 180 degrees. As can be seen, DeformingNeRF fails to handle twisting. Gaussian Frosting produces blurry artifacts across all angles. GaMeS shows spiky artifacts. While SuGaR works reasonably well at 90 degrees, it creates artifacts at larger angles. In contrast, our method remains stable and generates reasonable results even under extreme rotations.

\subsection{Other Control Methods}

\begin{figure*}
  \centering
  \includegraphics[width=\textwidth]{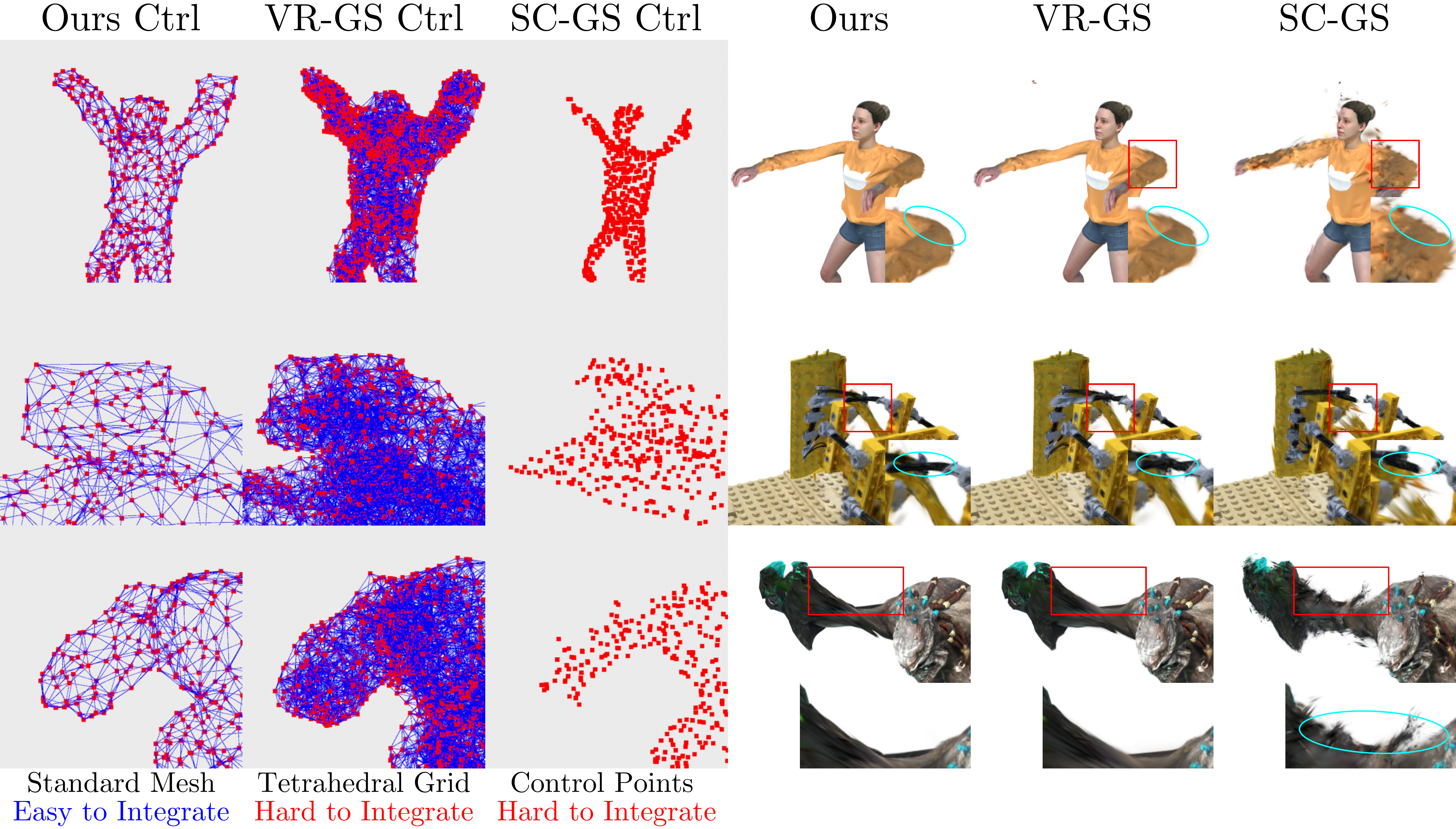}
  \caption{Comparison with methods that can be adapted for deformation. We show the method's control structures and their output. \mdefect{Cyan circles} mark defects, not having cyan circles indicates satisfactory results. Our method uses triangle mesh cages for control, which is simpler and easier to integrate with existing animation software such as Blender, unlike VR-GS or SC-GS. Our method also produces superior results. While VR-GS performed adequately on the mutant scene, it struggled with the lego scene's detailed deformations.}
  \label{fig:exp-scgs}
\end{figure*}

Our method controls 3DGS deformation using triangular mesh cages, unlike VR-GS \cite{vrgs}, which uses dense tetrahedral grids, and SC-GS \cite{SC-GS}, which uses sparse control points. Compared to these approaches, our triangular mesh approach is simpler and easier to integrate with existing 3D software or animation pipeline, and it achieves comparable, if not superior, quality in deformation. 

Since VR-GS and SC-GS are designed for different scenarios, we adapt them for deformation by directly manipulating their control structure. We evaluate these approaches against our method using three scenes from the D-NeRF \cite{dnerf} dataset: mutant, jumpingjacks, and lego. For fair comparison, we train SC-GS on these dynamic scenes and extract models at a single time frame for testing VR-GS and our method. This is because SC-GS requires training on video data, while other methods do not. We then analyze each method's control structure for controlling deformation and deformation results.

As shown in \Cref{fig:exp-scgs}, our method controls deformation using standard triangle meshes, while VR-GS uses dense tetrahedral grids and SC-GS uses sparse control points. Our approach offers a simpler structure than VR-GS's simulation-focused design, and integrates better with existing animation software than both SC-GS and VR-GS. In terms of result quality, our method demonstrates clear advantages over SC-GS, producing smoother shoulder deformations in the jumpingjacks and mutant scenes, while avoiding the disconnection artifacts present in the lego scene. When compared to VR-GS, our method produces smoother details on the lego and jumpingjacks scenes, and the results are comparable in the mutant scene. This shows that our method is more effective in handling complex deformations while using a simpler and easier-to-integrate control method.

\subsection{Training \& Deformation Speed}

We then benchmark the training and deformation times across all methods. We test on two sets of scenes/cages:

\textbf{NeRF Scenes/Cages} We select scenes from the NeRF Synthetic Dataset \cite{nerf} and use the cages from our cage construction algorithm for deformation. The selected scenes are: lego, chair, ficus, and hotdog.

\textbf{DeformingNeRF Scenes/Cages} We also test on the scenes selected by DeformingNeRF \cite{deforming-nerf}, using DeformingNeRF's cages as well. More concretely, DeformingNeRF \cite{deforming-nerf} selected two scenes from the NeRF Synthetic Dataset \cite{nerf}(chair and lego) and two from the NSVF \cite{nsvf} Synthetic Dataset (robot and toad).

We chose the ficus and hotdog scene over the robot and toad scene to test the method's performance when scaling objects with significant details or splitting Gaussians representing flat surfaces. In terms of cages, our cages are automatically generated and undergo extensive deformations, while DeformingNeRF's cages are manually created and have milder deformations.

\begin{table} 
    \centering
    \resizebox{\linewidth}{!}{
        \Huge
        \begin{tabular}{c||*{1}{c}||*{1}{c}}
        \toprule
        \multirow{2}{*}{Method} & \multicolumn{1}{c}{NeRF \cite{nerf} Scenes} 
        & \multicolumn{1}{c}{DeformingNeRF \cite{deforming-nerf} Scenes}   \\
        & training (sec$\downarrow$) &  training (sec$\downarrow$) \\
        \midrule
        DeformingNeRF \cite{deforming-nerf} & 491.33 & 479.18 \\
        SuGaR \cite{sugar} & 3234.30 & 3166.88 \\
        GaMeS \cite{games} & \fst{432.02} & \snd{469.65} \\
        Frosting \cite{GaussianFrosting} & 2649.78 & 2673.52 \\
        Vanilla 3DGS \cite{3dgs} & \snd{460.19} & \fst{462.38} \\
        \midrule
        Ours & \fst{N/A} & \fst{N/A} \\
        \bottomrule
    \end{tabular}
    }
    \vspace{0.1em}
   \caption{Benchmark results comparing training times. \fst{Red} indicates best values, \snd{blue} marks second-best. Note that with a pre-trained vanilla 3DGS model, our method can directly deform it without retraining or conversion.}
   \label{fig:exp-train-speed-benchmark}
\end{table}

\begin{table} 
    \centering
    \resizebox{\linewidth}{!}{
        \Huge
\begin{tabular}{cc|*{1}{c}||cc|*{1}{c}}
    \toprule
    
    \multicolumn{3}{c}{NeRF \cite{nerf} Scenes} & \multicolumn{3}{c}{DeformingNeRF \cite{deforming-nerf} Scenes}   \\
    \multirow{2}{*}{Scene} & \multirow{2}{*}{Method} & training & \multirow{2}{*}{Scene} & \multirow{2}{*}{Method} & training  \\
    & & (sec$\downarrow$) & & & (sec$\downarrow$) \\
    
    \midrule

    \midrule \multirow{6}{*}{chair} & DeformingNeRF \cite{deforming-nerf} & 451.15 & \multirow{6}{*}{chair} & DeformingNeRF \cite{deforming-nerf} & 451.15 \\ 
     & SuGaR \cite{sugar} & 3154.14 & & SuGaR \cite{sugar} & 3154.14 \\ 
     & GaMeS \cite{games} & \snd{427.86} & & GaMeS \cite{games} & \snd{427.86} \\ 
     & Frosting \cite{GaussianFrosting} & 2651.98 & & Frosting \cite{GaussianFrosting} & 2651.98\\ 
     & Vanilla 3DGS \cite{3dgs} & \fst{417.16} & & Vanilla 3DGS \cite{3dgs} & \fst{417.16} \\ 
     & Ours & \fst{N/A} & & Ours & \fst{N/A} \\

    \midrule \multirow{6}{*}{lego} & DeformingNeRF \cite{deforming-nerf} & 515.22 & \multirow{6}{*}{lego} & DeformingNeRF \cite{deforming-nerf} & 515.22  \\ 
     & SuGaR \cite{sugar} & 3264.54 & & SuGaR \cite{sugar} & 3264.54 \\ 
     & GaMeS \cite{games} & \snd{462.17} & & GaMeS \cite{games} & \snd{462.17} \\ 
     & Frosting \cite{GaussianFrosting} & 2726.72 & & Frosting \cite{GaussianFrosting} & 2726.72 \\ 
     & Vanilla 3DGS \cite{3dgs} & \fst{457.76} & & Vanilla 3DGS \cite{3dgs} & \fst{457.76} \\ 
     & Ours & \fst{N/A} & & Ours & \fst{N/A} \\ 
     
    \midrule \multirow{6}{*}{hotdog} & DeformingNeRF \cite{deforming-nerf} & 621.38 & \multirow{6}{*}{robot} & DeformingNeRF \cite{deforming-nerf} & 439.98 \\ 
     & SuGaR \cite{sugar} & 3475.71 & & SuGaR \cite{sugar} & 3060.61  \\ 
     & GaMeS \cite{games} & \fst{401.03} & & GaMeS \cite{games} & \snd{407.78} \\ 
     & Frosting \cite{GaussianFrosting} & 2606.45 & & Frosting \cite{GaussianFrosting} & 2603.48 \\ 
    & Vanilla 3DGS \cite{3dgs} & \snd{535.54} & & Vanilla 3DGS \cite{3dgs} & \fst{400.25} \\ 
    & Ours & \fst{N/A} & & Ours & \fst{N/A} \\ 

    \midrule \multirow{6}{*}{ficus} & DeformingNeRF \cite{deforming-nerf} & \fst{377.58} & \multirow{6}{*}{toad} & DeformingNeRF \cite{deforming-nerf} & \fst{510.36}   \\ 
     & SuGaR \cite{sugar} & 3042.79 & & SuGaR \cite{sugar} & 3188.21 \\ 
     & GaMeS \cite{games} & 437.01 & & GaMeS \cite{games} & 580.80  \\ 
     & Frosting \cite{GaussianFrosting} & 2613.99 & & Frosting \cite{GaussianFrosting} & 2711.88 \\ 
     & Vanilla 3DGS \cite{3dgs} & \snd{430.30} & & Vanilla 3DGS \cite{3dgs} & \snd{574.36} \\
    & Ours & \fst{N/A} & & Ours & \fst{N/A} \\ 

    \bottomrule
\end{tabular}
    }
    \vspace{0.1em}
   \caption{Detailed per-scene training time of \Cref{fig:exp-train-speed-benchmark}. \fst{Red} indicates best values, \snd{blue} marks second-best. Training times of vanilla 3DGS are also provided for reference.}
    \vspace{-2em}
   \label{fig:exp-train-speed-benchmark-per-scene}
\end{table}

\textbf{Training Speed} \Cref{fig:exp-train-speed-benchmark} presents the average training time for all methods, while the detailed per-scene results are shown in \Cref{fig:exp-train-speed-benchmark-per-scene}. Training times of vanilla 3DGS are provided for reference. With a pre-trained vanilla 3DGS model or its variants, our method can directly deform it without retraining or conversion, hence no training would be needed. In contrast, other approaches require retraining or conversion because they altered the architecture of 3DGS for editability.

\begin{table*} 
    \centering
    \resizebox{\textwidth}{!}{
    \Huge
        \begin{tabular}{c||*{3}{c}||*{3}{c}}
        \toprule
        & \multicolumn{3}{c}{NeRF \cite{nerf} Scenes (Automatic Cages)} 
        & \multicolumn{3}{c}{DeformingNeRF \cite{deforming-nerf} Scenes (Manual Cages)}   \\
        Method & preprocess & deform & render
        & preprocess & deform & render  \\
        & (ms$\downarrow$) & (ms$\downarrow$/FPS$\uparrow$) & (ms$\downarrow$/FPS$\uparrow$)
        & (ms$\downarrow$) & (ms$\downarrow$/FPS$\uparrow$) & (ms$\downarrow$/FPS$\uparrow$) \\
        \midrule
        DeformingNeRF$^*$ \cite{deforming-nerf} & 3530.10 & 3933.93 / 0.25FPS & 4970.13 / 0.20FPS & 2642.48 & 2420.32 / 0.41FPS & 3441.58 / 0.29FPS \\
        SuGaR \cite{sugar} & \snd{1197.03} & 1483.37 / 0.67FPS & 17.26 / 57.94FPS & \snd{746.24} & 1474.52 / 0.68FPS & 16.78 / 59.59FPS \\
        GaMeS \cite{games} & 1517.75 & \fst{8.34 / 119.90FPS} & \snd{5.87 / 170.36FPS} & 1183.13 & \fst{6.56 / 152.44FPS} & \snd{6.24 / 160.26FPS} \\
        Frosting \cite{GaussianFrosting} & \fst{1163.60} & 125.83 / 7.95FPS & 17.27 / 57.90FPS & \fst{715.36} & 122.48 / 8.16FPS & 16.62 / 60.17FPS \\
        \midrule
        Ours & 3565.11 & \snd{16.42 / 60.90FPS} & \fst{4.12 / 242.72FPS} & 2744.05 & \snd{12.33 / 81.10FPS} & \fst{3.97 / 251.89FPS} \\
        \bottomrule
    \end{tabular}
    }
    \vspace{-1em}
    \begin{flushleft}
        \footnotesize ${^*}$ DeformingNeRF performs deformation during rendering. Note that the render time involves the deformation time.
    \end{flushleft}
    \vspace{-0.5em}
   \caption{Benchmark results comparing deformation times. \fst{Red} indicates best values, \snd{blue} marks second-best. Deformation times include once-per-scene preprocessing and actual deformation. The time to render the deformed representation is also presented here. Our method achieves real-time performance ($\sim$60FPS) for both cage types and is the fastest in rendering.}
   \label{fig:exp-quant-benchmark}
   \vspace{-1em}
\end{table*}

\begin{table} 
    \centering
    \resizebox{\linewidth}{!}{
        \Huge
        \begin{tabular}{cc||*{3}{c}}
    \toprule & & \multicolumn{3}{c}{NeRF \cite{nerf} Scenes} \\
    \multirow{2}{*}{Scene} & \multirow{2}{*}{Method} & preprocess & deform & render  \\
    & & (ms$\downarrow$) & (ms$\downarrow$/FPS$\uparrow$) & (ms$\downarrow$/FPS$\uparrow$) \\
    \midrule
    
    \midrule \multirow{5}{*}{chair} & DeformingNeRF \cite{deforming-nerf} & 3361.99 & 2882.62 / 0.35FPS & 3908.45 / 0.26FPS \\ 
     & SuGaR \cite{sugar} & \snd{1055.86} & 1528.79 / 0.65FPS & 17.04 / 58.69FPS \\ 
     & GaMeS \cite{games} & 1355.31 & \fst{7.25 / 137.93FPS} & \snd{5.42 / 184.50FPS} \\ 
     & Frosting \cite{GaussianFrosting} & \fst{1023.09} & 125.62 / 7.96FPS & 16.36 / 61.12FPS \\ 
     & Ours & 3124.88 & \snd{13.89 / 71.99FPS} & \fst{3.43 / 291.55FPS} \\ 
          
    \midrule \multirow{5}{*}{lego} & DeformingNeRF \cite{deforming-nerf} & 4075.17 & 4633.86 / 0.22FPS & 5668.55 / 0.18FPS \\ 
     & SuGaR \cite{sugar} & \snd{1510.96} & 1511.84 / 0.66FPS & 16.91 / 59.14FPS \\ 
     & GaMeS \cite{games} & 2188.09 & \fst{12.09 / 82.71FPS} & \snd{6.43 / 155.52FPS} \\ 
     & Frosting \cite{GaussianFrosting} & \fst{1495.65} & 133.17 / 7.51FPS & 16.10 / 62.11FPS \\ 
     & Ours & 5148.76 & \snd{23.71 / 42.18FPS} & \fst{4.12 / 242.72FPS} \\ 
          
    \midrule \multirow{5}{*}{hotdog} & DeformingNeRF \cite{deforming-nerf} & 3500.92 & 3989.47 / 0.25FPS & 5028.07 / 0.20FPS \\ 
     & SuGaR \cite{sugar} & 1234.86 & 1477.10 / 0.68FPS & 17.46 / 57.27FPS \\ 
     & GaMeS \cite{games} & \fst{803.10} & \fst{4.43 / 225.73FPS} & \snd{4.30 / 232.56FPS} \\ 
     & Frosting \cite{GaussianFrosting} & \snd{1156.40} & 122.38 / 8.17FPS & 17.01 / 58.79FPS \\ 
     & Ours & 1918.72 & \snd{9.53 / 104.93FPS} & \fst{3.25 / 307.69FPS} \\ 

    \midrule \multirow{5}{*}{ficus} & DeformingNeRF \cite{deforming-nerf} & 3182.31 & 4229.77 / 0.24FPS & 5275.44 / 0.19FPS \\ 
     & SuGaR \cite{sugar} & \snd{986.44} & 1415.76 / 0.71FPS & 17.61 / 56.79FPS \\ 
     & GaMeS \cite{games} & 1724.51 & \fst{9.59 / 104.28FPS} & \snd{7.34 / 136.24FPS} \\ 
     & Frosting \cite{GaussianFrosting} & \fst{979.27} & 122.16 / 8.19FPS & 19.61 / 50.99FPS \\ 
     & Ours & 4068.09 & \snd{18.53 / 53.97FPS} & \fst{5.68 / 176.06FPS} \\ 

    \bottomrule
\end{tabular}
    }
    \vspace{0.1em}
   \caption{Detailed per-scene deformation time of \Cref{fig:exp-quant-benchmark} on NeRF scenes. Note that our method is consistently the second-fastest in deformation and fastest in rendering.}
   \label{tbl:deforming-nerf-scene-deform-benchmark-nerf}
\end{table}

\begin{table} 
    \centering
    \resizebox{\linewidth}{!}{
        \Huge
        \begin{tabular}{cc||*{3}{c}}

    \toprule & & \multicolumn{3}{c}{DeformingNeRF \cite{deforming-nerf} Scenes} \\
    
    \multirow{2}{*}{Scene} & \multirow{2}{*}{Method} & preprocess & deform & render  \\
    & & (ms$\downarrow$) & (ms$\downarrow$/FPS$\uparrow$) & (ms$\downarrow$/FPS$\uparrow$) \\
    
    \midrule
    
    \midrule \multirow{5}{*}{chair} & DeformingNeRF \cite{deforming-nerf} & 3422.40 & 2910.13 / 0.34FPS & 3936.68 / 0.25FPS \\ 
     & SuGaR \cite{sugar} & \snd{1061.50} & 1486.58 / 0.67FPS & 16.91 / 59.14FPS \\ 
     & GaMeS \cite{games} & 1353.42 & \fst{7.49 / 133.51FPS} & \snd{5.43 / 184.16FPS} \\ 
     & Frosting \cite{GaussianFrosting} & \fst{1018.24} & 126.04 / 7.93FPS & 16.98 / 58.89FPS \\ 
     & Ours & 3139.67 & \snd{14.31 / 69.88FPS} & \fst{3.71 / 269.54FPS} \\ 
     
    \midrule \multirow{5}{*}{lego} & DeformingNeRF \cite{deforming-nerf} & 2084.41 & 1197.90 / 0.83FPS & 2219.81 / 0.45FPS \\ 
     & SuGaR \cite{sugar} & \snd{441.99} & 1532.87 / 0.65FPS & 16.83 / 59.42FPS \\ 
     & GaMeS \cite{games} & 593.17 & \fst{3.36 / 297.62FPS} & \snd{6.17 / 162.07FPS} \\ 
     & Frosting \cite{GaussianFrosting} & \fst{424.85} & 125.03 / 8.00FPS & 15.84 / 63.13FPS \\ 
     & Ours & 1378.66 & \snd{6.18 / 161.81FPS} & \fst{3.94 / 253.81FPS} \\ 
     
    \midrule \multirow{5}{*}{robot} & DeformingNeRF \cite{deforming-nerf} & 1895.59 & 1441.66 / 0.69FPS & 2440.74 / 0.41FPS \\ 
     & SuGaR \cite{sugar} & \snd{517.81} & 1421.84 / 0.70FPS & 16.43 / 60.86FPS \\ 
     & GaMeS \cite{games} & 706.71 & \fst{3.86 / 259.07FPS} & \snd{5.68 / 176.06FPS} \\ 
     & Frosting \cite{GaussianFrosting} & \fst{510.92} & 117.94 / 8.48FPS & 17.00 / 58.82FPS \\ 
     & Ours & 1664.74 & \snd{7.19 / 139.08FPS} & \fst{3.94 / 253.81FPS} \\
     
    \midrule \multirow{5}{*}{toad} & DeformingNeRF \cite{deforming-nerf} & 3167.54 & 4131.60 / 0.24FPS & 5169.09 / 0.19FPS \\ 
     & SuGaR \cite{sugar} & \snd{963.66} & 1456.78 / 0.69FPS & 16.95 / 59.00FPS \\ 
     & GaMeS \cite{games} & 2079.23 & \fst{11.51 / 86.88FPS} & \snd{7.69 / 130.04FPS} \\ 
     & Frosting \cite{GaussianFrosting} & \fst{907.43} & 120.92 / 8.27FPS & 16.67 / 59.99FPS \\ 
     & Ours & 4793.12 & \snd{21.63 / 46.23FPS} & \fst{4.29 / 233.10FPS} \\ 
     
    \bottomrule
\end{tabular}
    }
    \vspace{0.1em}
   \caption{Detailed per-scene deformation time of \Cref{fig:exp-quant-benchmark} on DeformingNeRF scenes. Note that our method is consistently the second-fastest in deformation and fastest in rendering.}
   \label{tbl:deforming-nerf-scene-deform-benchmark-deformingnerf}
\end{table}

\textbf{Deformation Speed} \Cref{fig:exp-quant-benchmark} presents the average deformation time for all methods, and the detailed per-scene results are provided in \Cref{tbl:deforming-nerf-scene-deform-benchmark-nerf} and \Cref{tbl:deforming-nerf-scene-deform-benchmark-deformingnerf}. As \Cref{fig:exp-quant-benchmark} presents, our method, on average, achieves 60FPS on the simpler DeformingNeRF cages and our more challenging cages, hence real-time deformation. In terms of once-per-scene preprocessing, our approach is slower compared to mesh-based 3DGS methods like SuGaR \cite{sugar}, GaMeS \cite{games}, and Gaussian Frosting \cite{GaussianFrosting}. This is because our method needs to process more points for deformation and splitting, while mesh-based methods can simply deform the underlying mesh or triangle soup. However, our method is the fastest in rendering, as we use the unmodified vanilla 3DGS rendering process. 

This can also be seen from the per-scene results shown in \Cref{tbl:deforming-nerf-scene-deform-benchmark-nerf} and \Cref{tbl:deforming-nerf-scene-deform-benchmark-deformingnerf}. As can be seen, our method is consistently the second-fastest in deformation and fastest in rendering.

\section{Applications}

Our method can be extended to work with other methods or 3DGS variants, making it valuable for many application scenarios.

\subsection{Composition and Animation}

\begin{figure*}
  \centering
  \includegraphics[width=\linewidth]{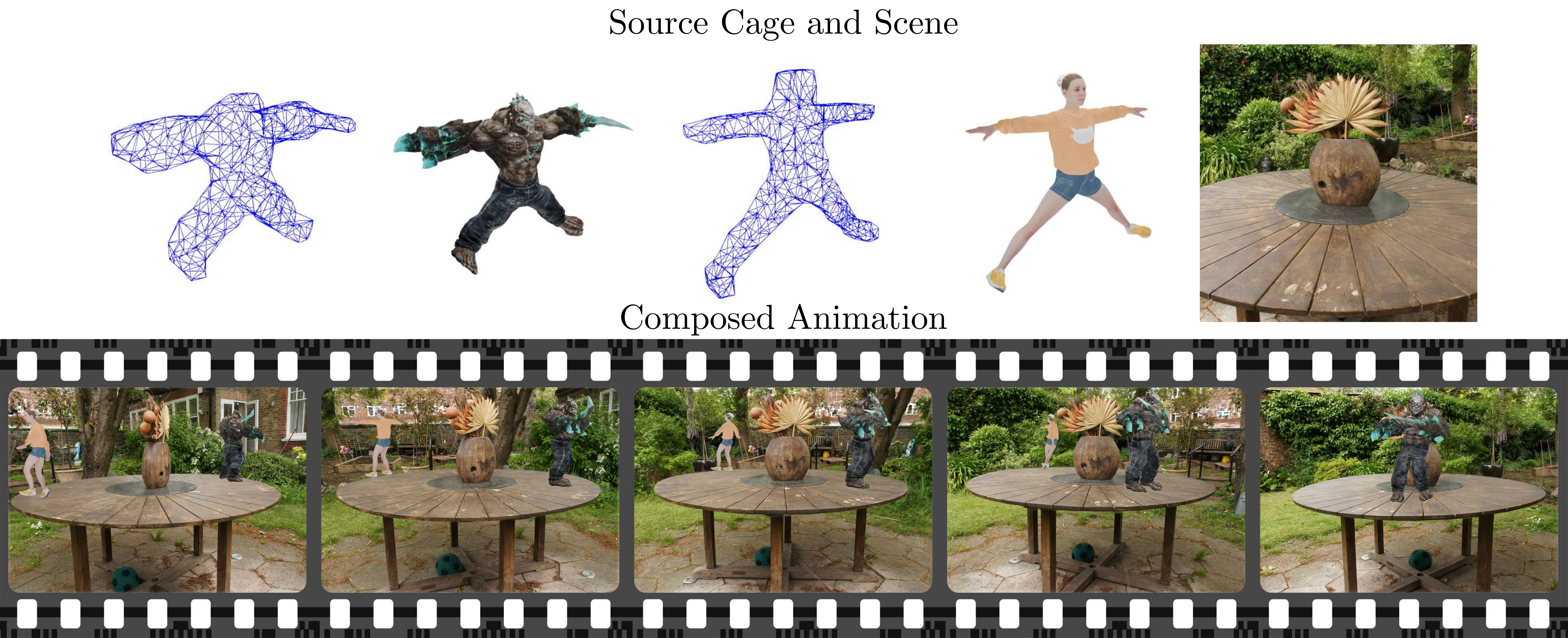}
  \caption{Animating and composing scenes using our method. Note that our cage-based deformation algorithm can be used for animation and composition by animating the cages and placing the cages into another scene. Please refer to our supplementary video for more details.}
  \label{fig:exp-animation}
\end{figure*}

Since our deformation algorithm enables cage-based deformation on 3DGS, we can animate and compose different 3DGS models by animating their cages and placing the cages into another scene. This animation and composition can be done using existing mesh editing and animation software, such as Blender \cite{blender}.

\Cref{fig:exp-animation} shows animated 3DGS versions of Mixamo \cite{mixamo} characters Sophie and Mutant, composited into the MipNeRF360 \cite{mipnerf360} garden scene. This is done by animating the cage of the characters using the automatic rigging and animation functionality of Mixamo, placing the cages within the garden scene, and applying our algorithm. As can be seen, our model achieves high-quality animation and composition of 3DGS models, which is useful for animators and artists intending to work with 3DGS.

\subsection{Combining with Editing Methods}

Our deformation algorithm achieves low-level, direct shape editing and can be integrated with other work on editing or editable 3DGS to complement their editing abilities. 

\begin{figure}
  \centering
  \includegraphics[width=\linewidth]{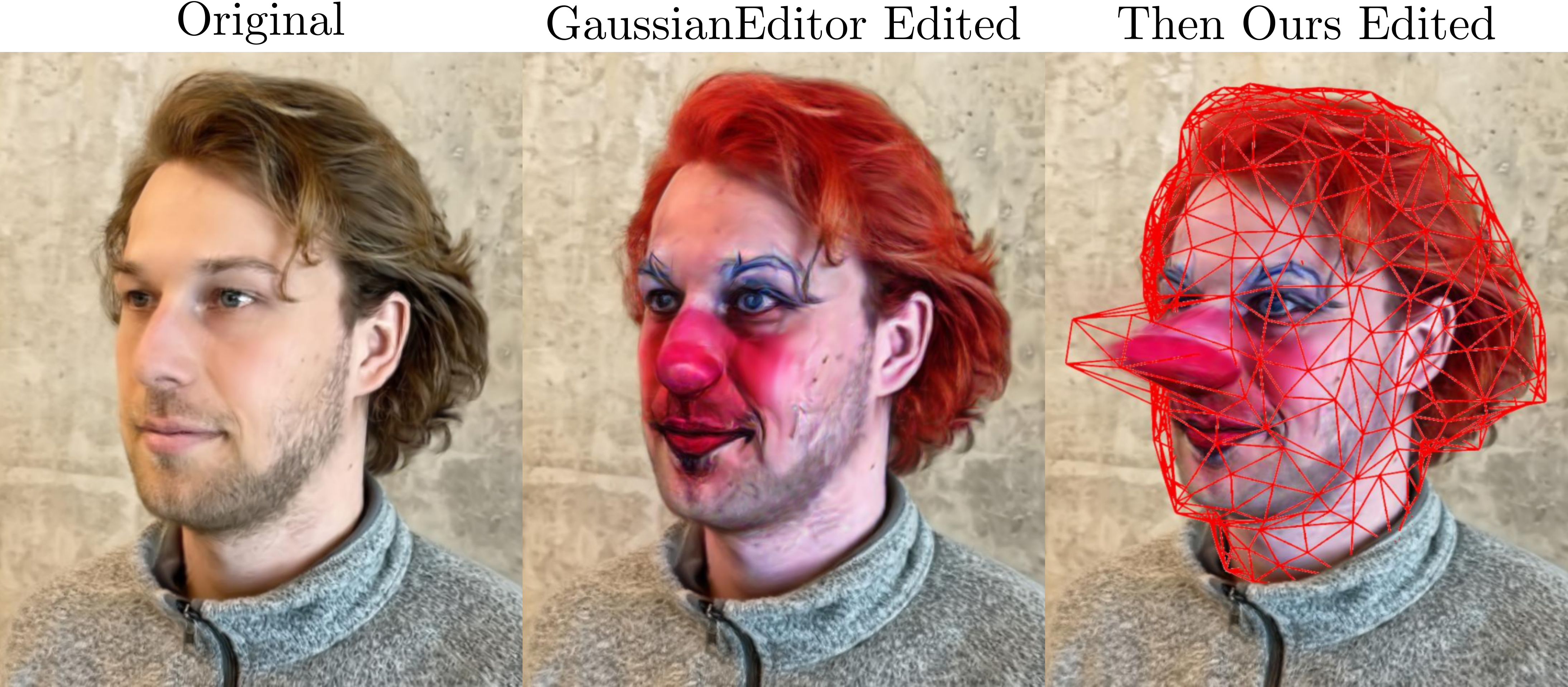}
  \caption{Integrating our method with GaussianEditor \cite{gaussianEditor}. Note that our approach successfully stretched the nose of the face edited by GaussianEditor. Please refer to our demo video for more results.}
  \label{fig:exp-int-gaussianeditor}
\end{figure}

\begin{figure}
  \centering
  \includegraphics[width=\linewidth]{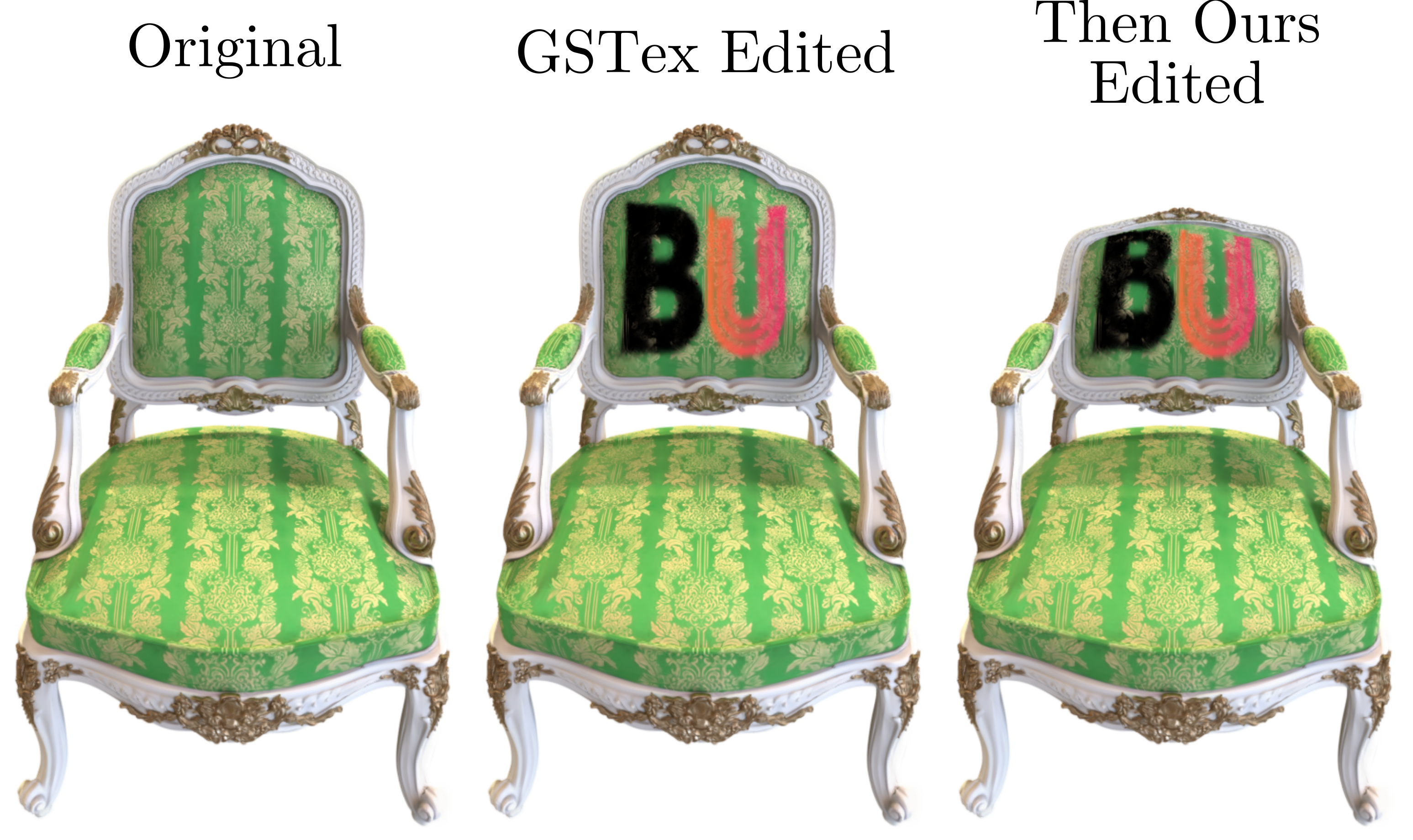}
  \caption{Integrating our method with GSTex \cite{gstex}. Note that GSTex enables texture editing, and our method enables shape editing combined to provide comprehensive editing capabilities.}
  \label{fig:exp-int-gstex}
\end{figure}

We start by integrating with GaussianEditor \cite{gaussianEditor}, a text-prompt-based 3DGS editing method. As shown in \Cref{fig:exp-int-gaussianeditor}, in the face scene from the Instruct-NeRF2NeRF Dataset \cite{in2n-nerf}, we use GaussianEditor to select the scene's facial region and edit its appearance with the text prompt "turn him into a clown". We then apply our deformation algorithm to stretch the nose of the edited face. As can be seen, our method successfully stretched the nose of the edited face.

Our method can also be integrated with GSTex\cite{gstex}, a 2DGS variant that allows texture editing. We tested this integration using a chair from the NeRF Synthetic Dataset\cite{nerf}. As shown in \Cref{fig:exp-int-gstex}, we modified the chair's texture using GSTex, then applied our deformation algorithm to deform the result. This combination demonstrates how our approach enhances the overall editing capabilities.

\subsection{Extending to Variants of 3DGS}

\begin{figure}
  \centering
  \includegraphics[width=\linewidth]{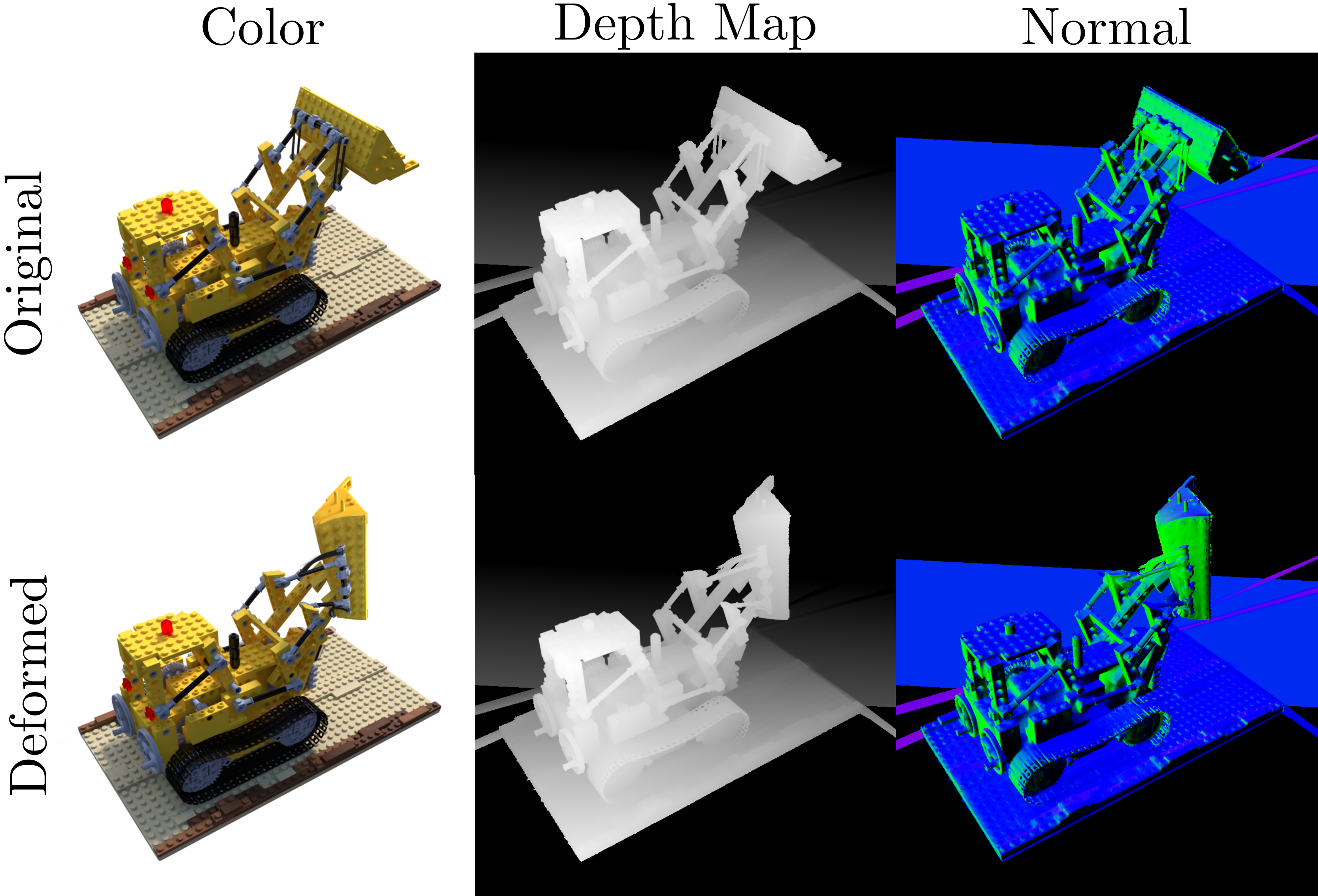}
  \caption{Integrating our method with 2DGS \cite{2dgs}. Note that the deformation performed by our method is not only high-quality in RGB rendering but also in depth and normal map as well.}
  \label{fig:exp-int-2dgs}
\end{figure}

\begin{figure}
  \centering
  \includegraphics[width=\linewidth]{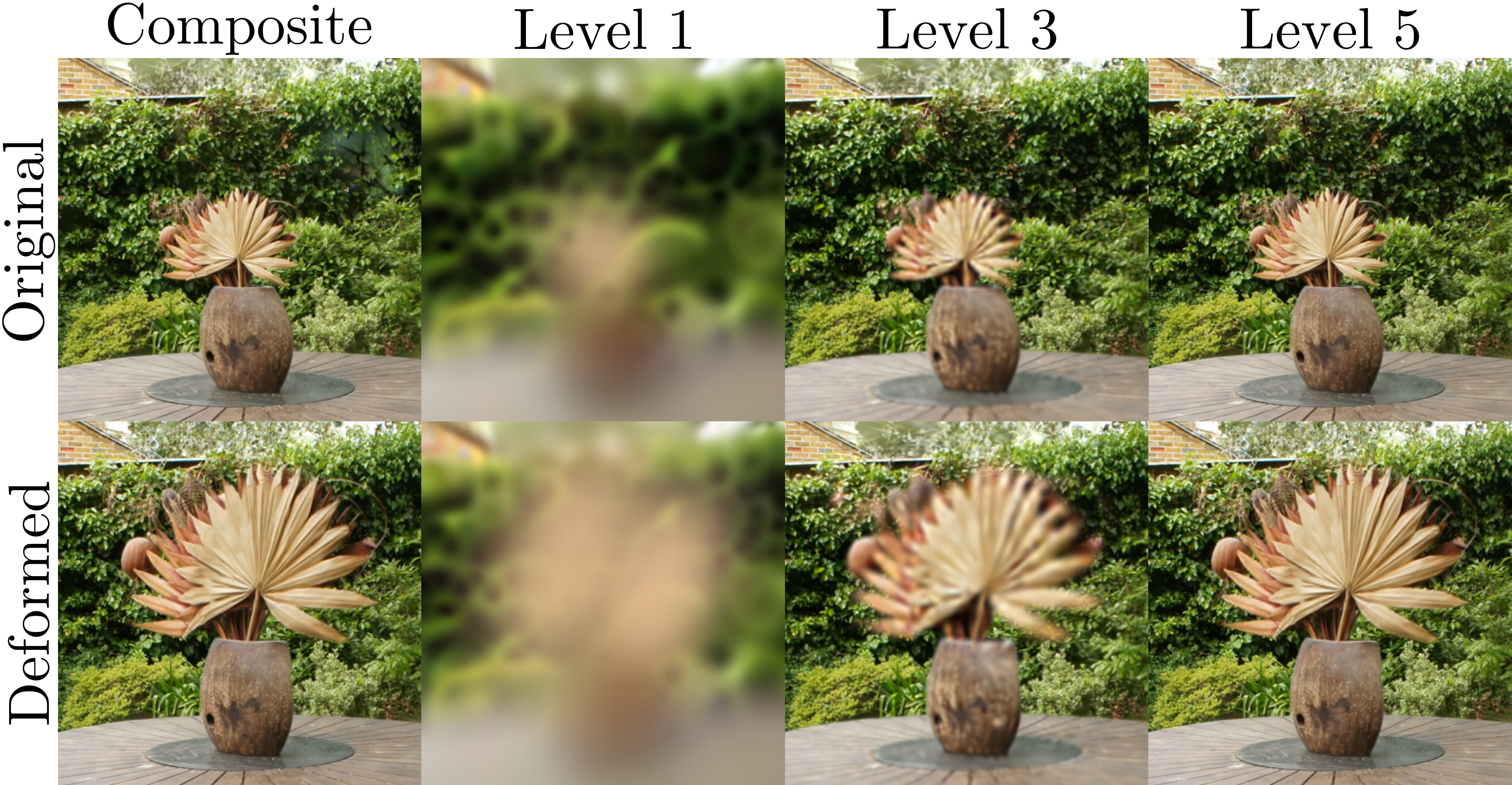}
  \caption{Integrating our method with FLoD \cite{flod}. Our method enlarged the flowers, which is correctly applied to all Level-of-Details(LoD) levels and the final composite result.}
  \label{fig:exp-int-flod}
\end{figure}

Our method can also be extended to work with other 3DGS variants.

We start by integrating with 2DGS \cite{2dgs}, a method that uses flattened 2D Gaussian disks instead of 3D Gaussian balls for improved geometry, depth rendering, and normal reconstruction. We demonstrate this by deforming a 2DGS capturing the lego scene from the NeRF Synthetic Dataset \cite{nerf}. Results are shown in \Cref{fig:exp-int-2dgs}; note the depth and normal render of the deformed model is high-quality as well.

We also integrate with FLoD \cite{flod}, a method that adds Level-of-Details(LoD) to 3DGS. We test this by editing a FLoD capturing the garden scene from the MipNeRF360 dataset \cite{mipnerf360}. As shown by \Cref{fig:exp-int-flod}, our deformation works well across all LoDs, demonstrating our method's adaptability to these variants of 3DGS. This adaptability could allow our method to utilize the available information across different models for different applications.

\subsection{Ablation Studies}

\begin{figure}
  \centering
  \includegraphics[width=\linewidth]{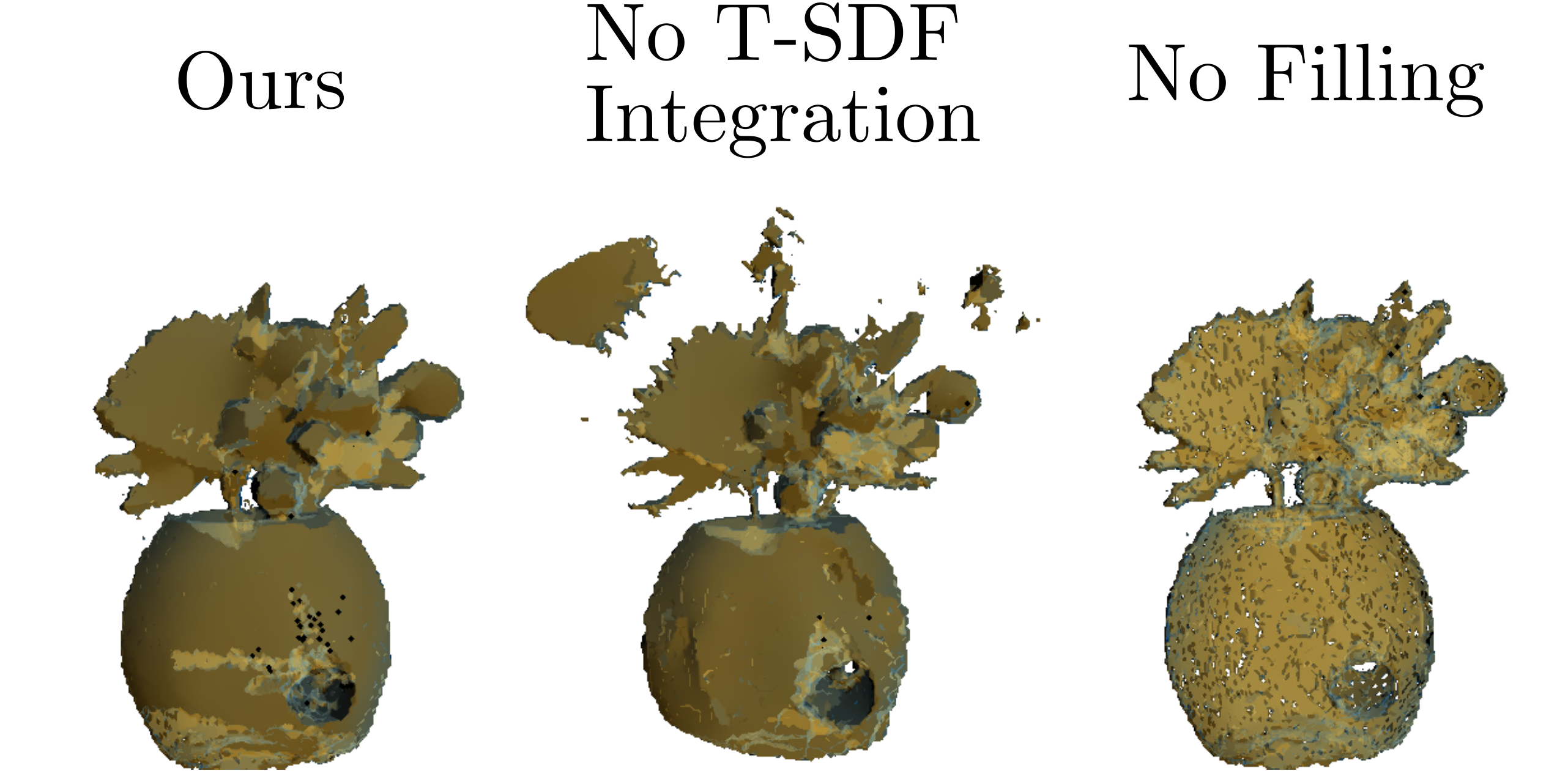}
  \caption{Ablation study results for cage building algorithm. The voxel grids before closing simplification are shown. Note that not performing T-SDF integration would lead to artifacts on top of the vase, while not performing filling leads to porous and hollow shapes. Both would harm cage quality.}
  \label{fig:exp-cage-ablation}
\end{figure}

\textbf{Cage-building Algorithm} We start our ablation study by analyzing the key steps in our cage-building algorithm, using the vase segmented from the garden scene in the MipNeRF360\cite{mipnerf360} dataset. We compare our voxel grid extraction method against two baselines: a naive approach that skips T-SDF integration and directly constructs voxel grids by performing depth carving using 3DGS-rendered depth maps ("No T-SDF Integration"), and a version of our algorithm that skips depth carving and use the surface points extracted from T-SDF integration instead ("No Filling").

\Cref{fig:exp-cage-ablation} shows the produced voxel grids prior to applying the closing operator. Using depth maps directly for voxel carving without T-SDF integration creates noise on the vase's top. Without filling, the resulting voxel grid is porous and hollow, which closing operations cannot effectively remedy. Both approaches produce suboptimal results and could harm the cage's quality.

\begin{figure}
  \centering
  \includegraphics[width=\linewidth]{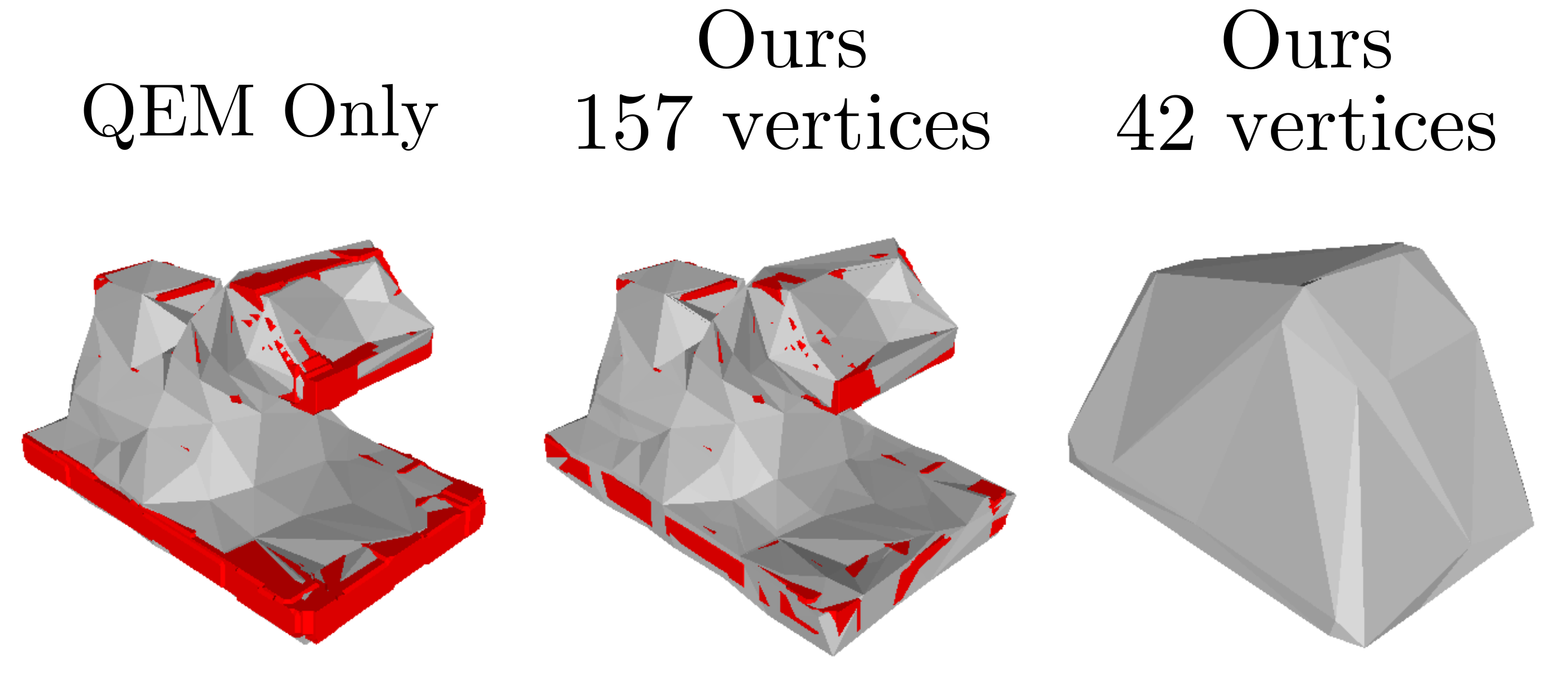} 
  \caption{Ablation study results for our two-stage mesh simplification process. Note how our two-stage method produces a cage (gray) that better envelops the object geometry (red) compared to the standard QEM approach. Further simplification perfects the enclosure but at the cost of detail, demonstrating a trade-off.}
  \label{fig:exp-ablation-two-stage}
\end{figure}

\begin{table}
  \centering
  \begin{tabular}{lcc}
    \toprule
     & Neg. MVC Entry & Smallest Neg. MVC Value \\
    \midrule
    Improvement & $8.7\%$ & $99.67\%$ \\
    \bottomrule
  \end{tabular}
    
  \vspace{0.5em}
  \caption{MVC statistics change for the lego scene. Our two-stage method makes 8.7\% of MVC entries non-negative and reduces the magnitude of the smallest negative weight by 99.67\%.}
  \label{tbl:neg-mvc-quant}
  \vspace{-3.5em}
\end{table}

\textbf{Two-Stage Mesh Simplification} We now validate our two-stage mesh simplification design, which is critical for generating high-quality cages. We compare our full algorithm against a baseline using only the first stage, namely Constrained QEM, on the lego scene from the NeRF Synthetic dataset \cite{nerf}.

We further quantify this comparison on the lego scene. Using the centers of Gaussian ellipsoids as geometry proxies, we evaluate their MVC weights on the cage before and after decimation. We then measure the percentage point drop in negative MVC entries and the relative change in the smallest negative (worst-case) MVC weight, reporting both in \Cref{tbl:neg-mvc-quant}.

The results are visualized in \Cref{fig:exp-ablation-two-stage}. As can be seen, the baseline QEM Only approach produces a cage that encloses the object too tightly, leaving significant portions of the dense mesh (shown in red) outside the cage. In contrast, our two-stage method (middle, 157 vertices) generates a cage that provides superior coverage. The inclusion of the MVC penalty term encourages the cage to properly envelop the object, while also decreasing the occurrence of negative MVC weights for interior points. This confirms that our two-stage design effectively produces cages that are structurally optimized for high-quality, stable deformations. While further simplifying the cage (right) eliminates unenclosed regions entirely and further reduces negative MVC weights, it causes a significant loss of detail, highlighting a trade-off between cage quality for deformation and its fidelity to the object's original shape. 

The effectiveness of our design can also be seen from \Cref{tbl:neg-mvc-quant}. Our method makes 8.7\% of MVC entries non-negative and reduces the magnitude of the smallest negative MVC value by 99.67\%. Showcasing its ability to significantly drive negative MVC weights toward zero.

\begin{figure}
  \centering
  \includegraphics[width=\linewidth]{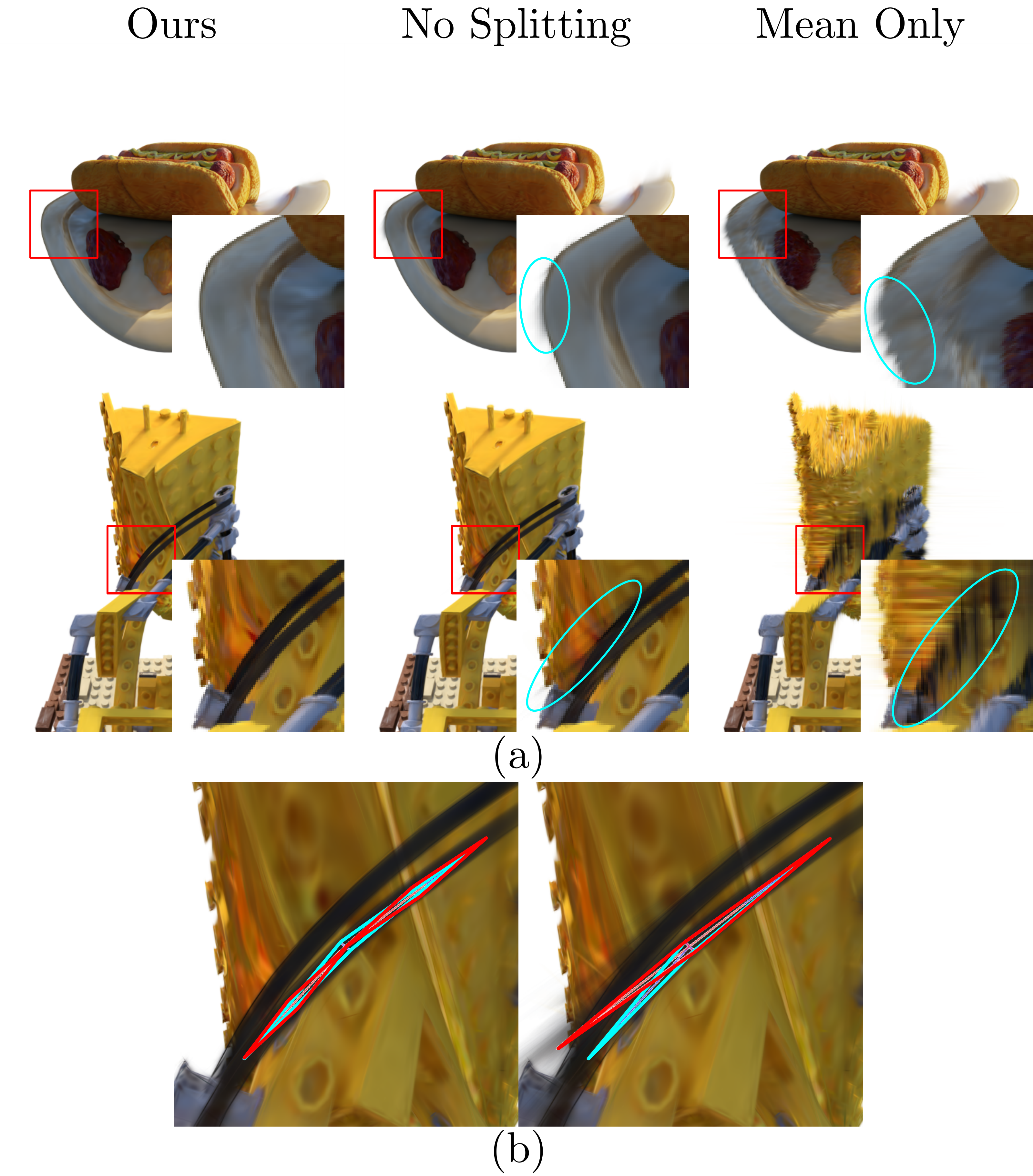}
  \caption{(a) Ablation study results for deformation algorithm. Note the sharp spikes caused by disabling splitting in the highlighted area. The naive mean-only variant produces significant artifacts as well. (b) The zoomed-in view of (a) highlights the deformed Gaussian (\textbf{\color{cyan}cyan diamond}) and the actual Gaussians (\textbf{\color{red}red diamond}, including both splitting and non-splitting Gaussians). Note that when the deformed Gaussian is approximated using only one Gaussian, rather than splitting, noticeable spiking artifacts appear.
 }
  \label{fig:exp-ablation}
\end{figure}



\textbf{Deformation Algorithm} To evaluate our design choices in the deformation algorithm, we compare our algorithm with two simpler variants: one without splitting and another that directly applies cage-based deformation to the position(mean) of the gaussians. We test these algorithms on the lego and hotdog scenes from the NeRF Synthetic Dataset \cite{nerf}. \Cref{fig:exp-ablation} shows the results.

As can be seen, the simpler mean-only variant of our algorithm produced significant striping artifacts. This is prevented in our algorithm by transforming the covariance (thus, rotation and shape) of the Gaussians as well. Furthermore, as shown in the highlighted and zoomed-in regions, using our algorithm without splitting results in spiking artifacts in bent regions. This occurs because the long belt is represented by a few thin, elongated ellipsoids. When bent, noticeable spiking artifacts appear, as shown in \Cref{fig:exp-ablation} (b). Clearly, bending is challenging to approximate using a single Gaussian ellipsoid instead of splitting. This demonstrates that the splitting procedure is essential for handling large deformations. Simply optimizing the existing Gaussians, as proposed by ARAP-GS \cite{arapgs} or VR-GS \cite{vrgs}, is insufficient for accurately modeling bending.

\begin{figure*}
  \centering
  \includegraphics[width=\linewidth]{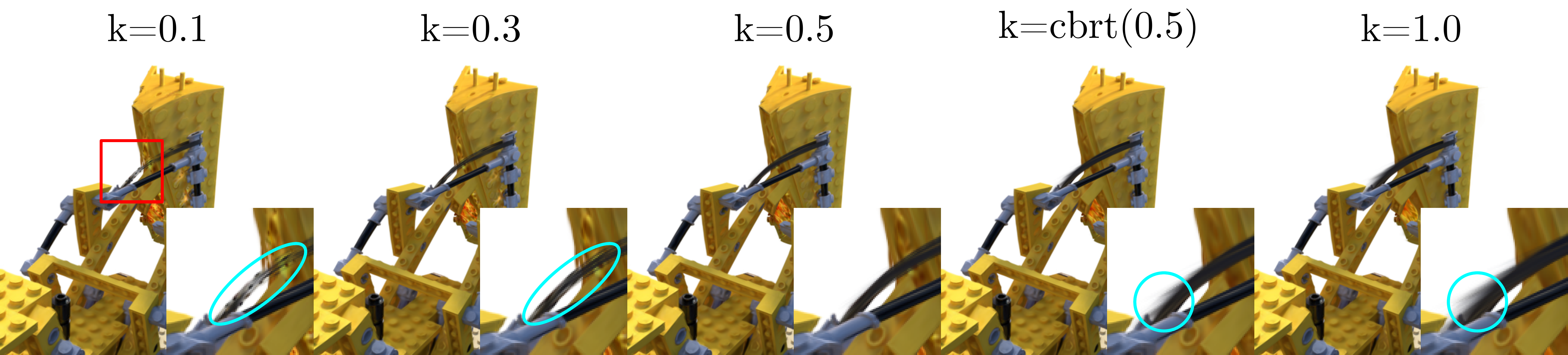}
  \vspace{-2em}
  \caption{Ablation study results for the split factor $k$. Note that setting $k$ too low creates fragmented surfaces, while setting it too high results in sharp spikes. It is shown that $0.5$ achieves a better balance than our theoretically derived value.}
  \label{fig:exp-ablation-factor}
  \vspace{-2em}
\end{figure*}

\textbf{Split Scaling Factor} Finally, we analyze our choice of scaling factor $k$ in the splitting step using the Lego scene from the NeRF Synthetic Dataset \cite{nerf}.

As shown in \Cref{fig:exp-ablation-factor}, the scaling factor $k$ significantly impacts results. A large $k$ (e.g., 1.0) creates spikes on the bulldozer's belt due to oversized split Gaussians failing to model the intricate bending deformation. Conversely, a small $k$ (e.g., 0.1) produces fragmented surfaces as the split Gaussians become too small to cover the belt. While our theoretical analysis suggests $k$ should be $\sqrt[3]{\frac{1}{2}}$ (denoted as cbrt(0.5)), empirically we find $\frac{1}{2}$ produces superior results.

\section{Conclusion}

In this paper, we introduced GSDeformer, a cage-based deformation algorithm for 3D Gaussian Splatting (3DGS). Our approach can directly deform existing trained vanilla 3DGS in real time and can be easily extended to its variants. We adapt cage-based deformation for 3DGS by first building a proxy point cloud from the Gaussians and then transferring the point cloud's deformation back to 3DGS, splitting the relevant Gaussians to handle bending. This approach requires no additional training data or architectural changes. We also developed an algorithm that automatically constructs cages for 3DGS deformation, featuring a two-stage simplification process that encourages the cage to better envelope the object and be structured for stable high-quality deformations.

\textbf{Limitations} Currently, our algorithm simply copies the spherical harmonics parameters for viewpoint-dependent color, without accounting for the effect of rotation. Additionally, compared to cage-based deformation, point-based deformation is better suited for creating smaller deformations, such as facial expressions. Extending 3DGS to capture and edit facial microexpressions is part of our future work.

%

\bibliographystyle{IEEEtran}
\bibliography{main}











\section{Biography Section}

\vspace{-33pt}

\begin{IEEEbiography}[{\includegraphics[width=1in,height=1.25in,clip,keepaspectratio]{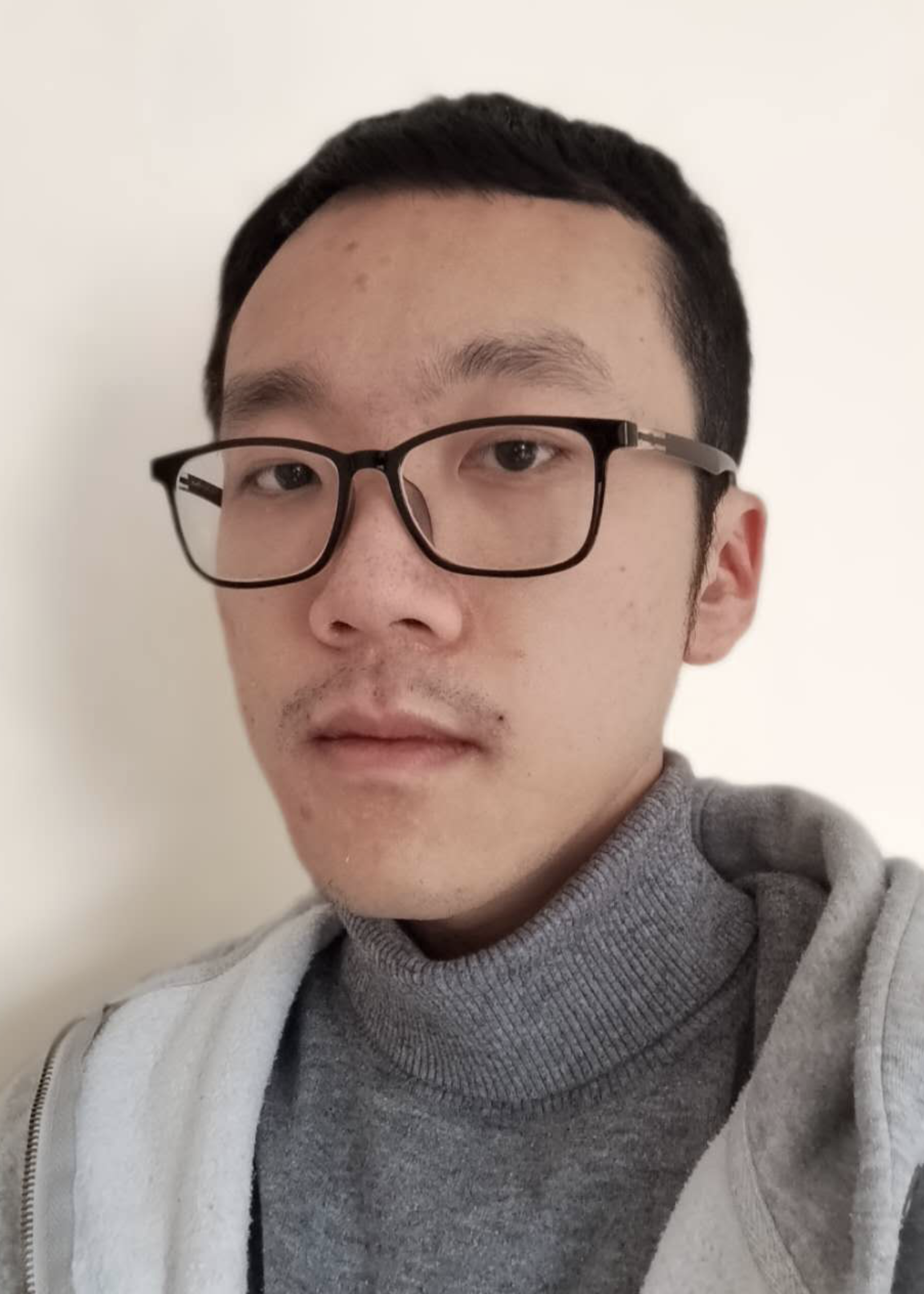}}]{Jiajun Huang}
Jiajun Huang is a Ph.D student at the National Centre for Computer Animation (NCCA) of Bournemouth University. He obtained his bachelor's degree from South China Normal University. His research interest covers neural scene representation understanding and editing. Currently, he is conducting research on 3D Gaussian Splatting understanding and animation.
\end{IEEEbiography}

\vspace{5pt}
\vspace{-33pt}

\begin{IEEEbiography}[{\includegraphics[width=1in,height=1.25in,clip,keepaspectratio]{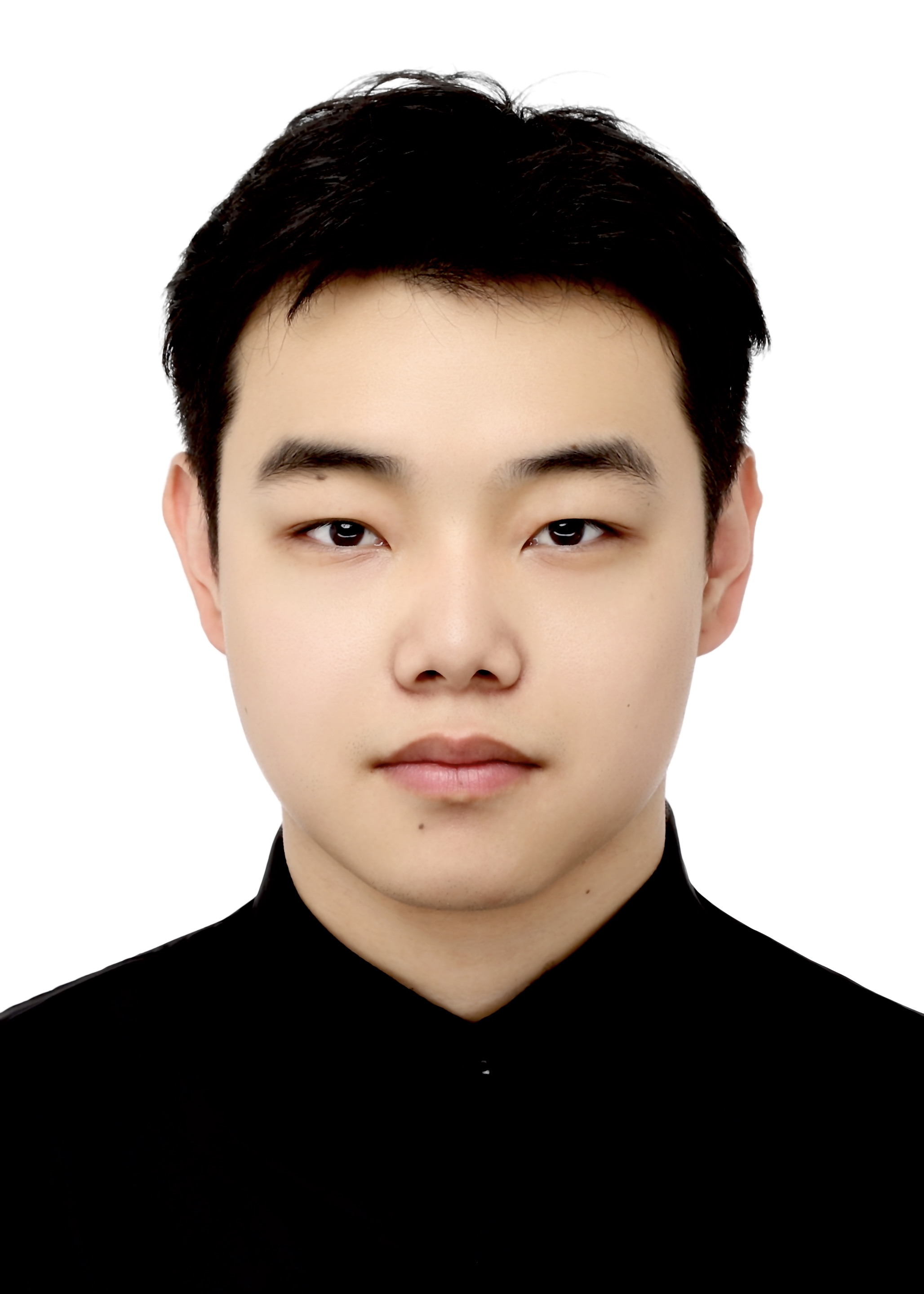}}]{Shuolin Xu}
Shuolin Xu is a Ph.D student at the National Centre for Computer Animation (NCCA) of Bournemouth University. He obtained his bachelor's degree from Zhengzhou University and his master's degree from Bournemouth University. Currently, he is conducting research on generative models, particularly in the areas of video generation and motion generation.
\end{IEEEbiography}

\vspace{5pt}
\vspace{-33pt}

\begin{IEEEbiography}[{\includegraphics[width=1in,height=1.25in,clip,keepaspectratio]{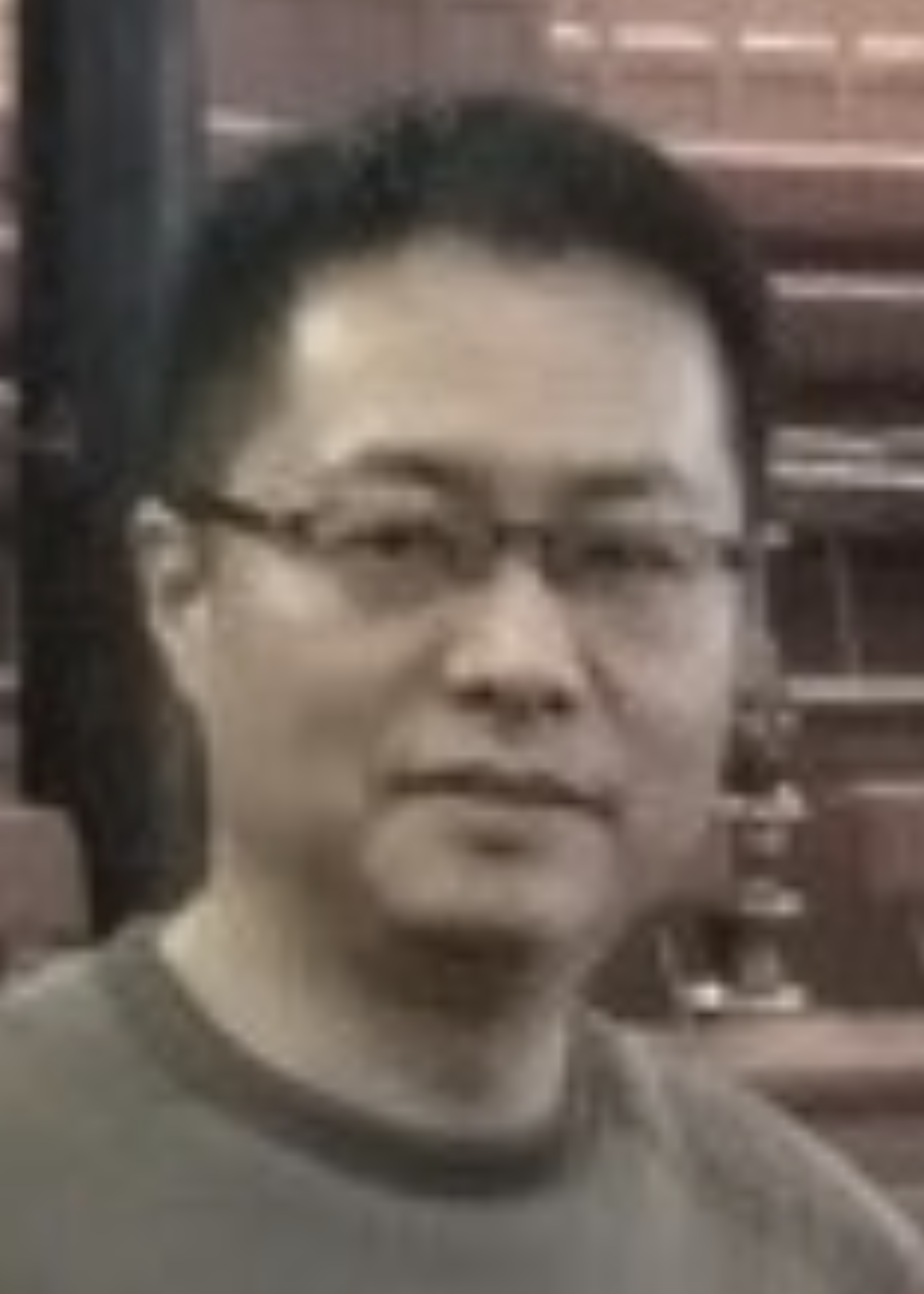}}]{Hongchuan Yu}
Hongchuan Yu is a Principal Academic of computer graphics in National Centre for Computer Animation, Bournemouth University, UK. He has published around 110 academic articles in reputable journals and conferences, and regularly served as PC members/referees for international journals and conferences. He is a Member of IEEE (MIEEE) and a fellow of High Education of Academy United Kingdom (FHEA). His research interests include Geometry, GenAI, Graphics, Image, and Video processing.
\end{IEEEbiography}

\vspace{5pt}
\vspace{-33pt}

\begin{IEEEbiography}[{\includegraphics[width=1in,height=1.25in,clip,keepaspectratio]{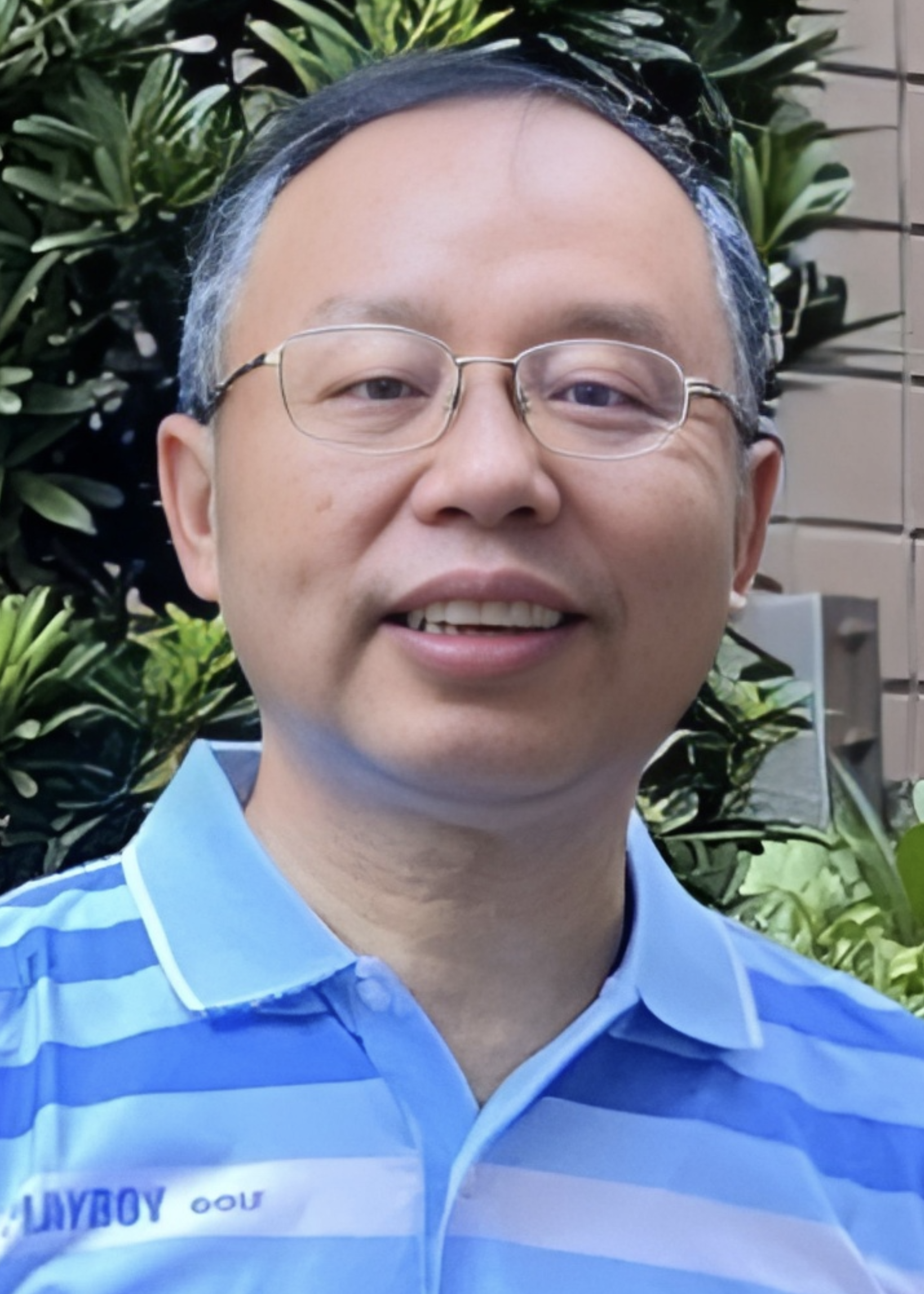}}]{Tong-Yee Lee}
Tong-Yee Lee (Senior Member, IEEE) received the Ph.D. degree in computer engineering from Washington State University, Pullman, in 1995. He is currently a chair professor with the Department of Computer Science and Information Engineering, National Cheng-Kung University (NCKU), Tainan, Taiwan. He leads the Computer Graphics Laboratory, NCKU (http://graphics.csie.ncku.edu.tw). His current research interests include computer graphics, nonphotorealistic rendering, medical visualization, virtual reality, and media resizing. He is a Senior Member of the IEEE and a Member of the ACM. He also serves on the editorial boards of both the IEEE Transactions on Visualization and Computer Graphics and IEEE Computer Graphics and Applications.
\end{IEEEbiography}



\newpage



\clearpage

\appendices

\section{Pseudo-code of Cage-building Algorithm}

The pseudo code of the cage-building algorithm is presented in \Cref{alg:cage-building} for reference.

\label{sect:apptx-pseudocode-cage}

\begin{algorithm}
\caption{Cage-Building Algorithm}
\label{alg:cage-building}

\begin{algorithmic}

\Require 3DGS Scene $S$, config parameters (num\_rings $n_r$, cameras\_per\_ring $n_c$, expand factor $s$)
\Ensure Simplified Cage Mesh $C$

\LComment{compute expanded bounding sphere}
\State $M$ = means of Gaussians in $S$
\State $m$ = Mean($M$)
\State $r$ = Max(L2 distance from points in $M$ to $m$) * $s$

\LComment{generate cameras surrounding $S$, with top and bottom}
\State $A$ = Generate cameras in $n_r$ evenly spaced rings of $n_c$ cameras each on a sphere of radius $r$.
\State Add cameras to $A$ at the top and bottom poles.
\State Orient all cameras in $A$ to look towards the center $m$.

\LComment{render depth image}
\State $D$ = Render3DGS($S$, $A$)

\LComment{reconstruct voxel grid}
\State $F_{tsdf}$ = TSDFIntegration($D$, $A$)
\State $V_s$ = ExtractSurfaceVoxel($F_{tsdf}$)
\State $D_{clean}$ = RenderTSDF($F_{tsdf}$, $A$)
\State $V_i$ = DepthCarving($D_{clean}$, $A$)
\State $V$ = merge voxels in $V_s$ and $V_i$

\LComment{smooth and simplify}
\State $V_b$ = MorphoClosing($V$)
\State $C_{raw}$ = MarchingCube($V_b$)
\State $C_{s}$ = BilateralFilter($C_{raw}$)
\State $C$ = EdgeCollapse($C_{s}$)

\Return $C$

\end{algorithmic}
\end{algorithm}

\section{Pseudo-code of Deformation Algorithm}

\label{sect:apptx-pseudocode}

\begin{algorithm}
\caption{GSDeformer Deformation Algorithm}
\label{alg:deformation}
\begin{algorithmic}

\Require 3DGS scene $S_s$, Source Cage $C_s$, Target Cage $C_d$, split threshold angle $t$.
\Ensure Deformed 3DGS scene $S_d$

\State ret = Empty3DGS()
\For{Gaussian $i$ in $GS$}
    \LComment{convert Gaussian to points-represented ellipsoids}
    \State $AP_s$ = point set of $i$ using \Cref{eq:point-set}
    
    \LComment{deform points using cage-based deformation}
    \State $AP_{mvc}$ = \text{EulerToMVC}($AP_s$, $C_s$)
    \State $AP_d$ = \text{MVCToEuler}($AP_{mvc}$, $C_d$)
    
    \LComment{perform splitting}
    \State $AP_{ss}$ = $[AP_s]$
    \State $AP_{ds}$ = $[AP_d]$
    \State $AP_{ss}$, $AP_{ds}$ = Split($i$, $AP_{ss}$, $AP_{ds}$, $t$)
    
    \LComment{transform Gaussians}
    \For {pre/post transform pair $s_i$, $d_i$ in $AP_{ss}$ and $AP_{ds}$}
        \State $\nmtx{T}$ = transform from $s_i$ to $d_i$ using \Cref{eq:estimate-transform}
        \State $\nvec{t}$ = $\nvec{c}$ in $d_i$ - $\nvec{c}$ in $s_i$
        \State $i'$ = transform $i$ using \Cref{eq:apply-transform} and (\ref{eq:apply-transform-1})
        \State append $i'$ to ret
    \EndFor
\EndFor
    
\State \Return ret

\end{algorithmic}
\end{algorithm}

\begin{algorithm}
\caption{Split Function}
\label{alg:split}
\begin{algorithmic}
\Require Gaussian $i$, source points $AP_s$, deformed points $AP_d$, split threshold angle $t$.
\Ensure split source points $AP_s$ and deformed points $AP_d$

\For{axis $a$ in $\{x,y,z\}$}
    \For{deformed ellipsoid $d_i$ in $AP_d$}
        \State $c$ = $d_i$'s center
        \State $l,r$ = endpoints of $d_i$'s $a$ axis 
        \State $\alpha$ = angle formed by $l$-$c$-$r$
        \If{$\alpha < t$}
            \LComment{Compute split ellipsoid 1}
            \State $d_l$ = copy of $d_i$
            \State center of $d_l$ = mean($c$, $l$)
            \State shift rest of $d_l$ points from $c$ to new center
            \State make $c$-$l$ $d_l$'s new $a$ axis
            \LComment{Compute split ellipsoid 2}
            \State $d_r$ = copy of $d_i$
            \State center of $d_r$ = mean($c$, $r$)
            \State shift rest of $d_r$ points from $c$ to new center
            \State make $c$-$r$ $d_r$'s new $a$ axis
            \LComment{update ellipsoid sets}
            \State replace $d_i$ in $AP_d$ using two ellipsoids $d_l$, $d_r$
            \State duplicate $d_i$'s source ellipsoid in $AP_s$
        \EndIf
    \EndFor
\EndFor

\State \Return $AP_s$, $AP_d$

\end{algorithmic}
\end{algorithm}

Our deformation algorithm is presented in \Cref{alg:deformation}. The split function is presented in \Cref{alg:split} . Additionally, it employs EulerToMVC() and MVCToEuler() from cage-based deformation to convert between Euclidean and cage-based coordinates for deformation.

To achieve real-time performance, we perform extensive caching in our implementation. Note that $AP_{mvc}$ and part of $T$ in \Cref{alg:deformation} can be precomputed. We also present a simplified implementation of the splitting process that directly operates on the deformed points-represented ellipsoids.

\section{Details of Cage-building Baseline}

\label{sect:cage-building-baseline}

Inspired by the 3DGS marching cube method proposed by SuGaR \cite{sugar}, our marching cube baseline creates a cage by first converting 3DGS into a binary occupancy voxel grid through opacity thresholding. The process then applies marching cubes to create meshes and simplify them through edge collapse, as proposed by NeRFShop \cite{nerf-shop}.

We transform 3DGS into a binary voxel grid by thresholding each voxel's opacity. A voxel becomes one if its opacity exceeds the threshold and zero otherwise. We calculate voxel opacity $d(\nvec{v})$ by summing contributions from the K-nearest 3D Gaussians from the voxel center:
\begin{equation}
d(\nvec{v}) = \sum_{g} \alpha_g \exp\left(-\frac{1}{2}(\nvec{v} - \nvec{\mu}_g)^\mathsf{T}\! \nmtx{\Sigma}^{-1}_g (\nvec{v} - \nvec{\mu}_g)\right)
\label{eq:gaussian_splatting_density}
\end{equation}

For each Gaussian $g$ near voxel center $\nvec{v}$, we use its opacity $\alpha_g$, mean $\nvec{\mu}_g$, and covariance $\nmtx{\Sigma}_g$. We use a threshold of 1e-6 and K=16 nearest Gaussians.

With the occupancy voxel grid, we smooth it using Bounding Proxy \cite{bounding-proxy}, mesh it using marching cubes, and simplify it using edge collapse to create the final cage. 

This cage-building method is later extended to work with 2DGS \cite{2dgs} by converting 2DGS to 3DGS: For every flat ellipse in 2DGS, we compute its third axis as the cross product of the first two axes and set its length to 1e-5, converting 2D ellipses into 3D ellipsoids.

\vfill

\end{document}